\documentclass[11pt]{article}

\usepackage[final]{acl}

\usepackage{times}
\usepackage{latexsym}

\usepackage[T1]{fontenc}

\usepackage[utf8]{inputenc}

\usepackage{microtype}

\usepackage{inconsolata}

\usepackage{graphicx}

\usepackage{graphicx}
\usepackage{caption}
\usepackage{amssymb}
\usepackage{amsmath}
\usepackage{multirow}
\usepackage{multicol}
\usepackage{booktabs}
\usepackage{float}
\usepackage{placeins}
\usepackage{xcolor}
\usepackage{colortbl}
\usepackage{ragged2e}
\usepackage{listings}
\definecolor{myred}{RGB}{178, 34, 34} 
\definecolor{mygreen}{RGB}{34,139,34}   
\definecolor{myred2}{RGB}{237, 211, 210} 
\definecolor{mygreen2}{RGB}{198, 232, 206} 
\definecolor{myblue2}{RGB}{218,232,252}

\definecolor{codegreen}{rgb}{0,0.6,0}
\definecolor{codegray}{rgb}{0.5,0.5,0.5}
\definecolor{codepink}{RGB}{252, 142, 172}
\definecolor{codepurple}{rgb}{0.58,0,0.82}
\definecolor{backcolour}{RGB}{245,245,245}

\lstdefinestyle{mystyle}{
    backgroundcolor=\color{backcolour},   
    commentstyle=\color{magenta},
    keywordstyle=\color{blue},
    numberstyle=\tiny\color{codegray},
    stringstyle=\color{codepurple},
    basicstyle=\fontfamily{\ttdefault}\footnotesize,
    breakatwhitespace=false,         
    breaklines=true,                 
    keepspaces=true,    
    frame=single,
    numbersep=5pt,                  
    showspaces=false,                
    showstringspaces=false,
    showtabs=false,                  
    tabsize=2,
    classoffset=1, 
    keywordstyle=\color{violet},
    classoffset=0,
}
\lstdefinelanguage{JavaScript}{
  keywords={typeof, new, true, false, catch, function, return, null, catch, switch, var, if, in, while, do, else, case, break},
  keywordstyle=\color{blue}\bfseries,
  ndkeywords={class, export, boolean, throw, implements, import, this},
  ndkeywordstyle=\color{darkgray}\bfseries,
  identifierstyle=\color{black},
  sensitive=false,
  comment=[l]{//},
  morecomment=[s]{/*}{*/},
  commentstyle=\color{purple}\ttfamily,
  stringstyle=\color{red}\ttfamily,
  morestring=[b]',
  morestring=[b]"
}
\usepackage{enumitem}\newcommand{\modelname}{DataSage}

\title{{\modelname}: Multi-agent Collaboration for Insight Discovery with External Knowledge Retrieval, Multi-role Debating, and Multi-path Reasoning}

\author{First Author \\
  Affiliation / Address line 1 \\
  Affiliation / Address line 2 \\
  Affiliation / Address line 3 \\
  \texttt{email@domain} \\\And
  Second Author \\
  Affiliation / Address line 1 \\
  Affiliation / Address line 2 \\
  Affiliation / Address line 3 \\
  \texttt{email@domain} \\}

\author{
Xiaochuan Liu\textsuperscript{1},
Yuanfeng Song\textsuperscript{1},
Xiaoming Yin\textsuperscript{1},
Xing Chen\textsuperscript{1} \\
\textsuperscript{1}ByteDance, China \\
}

\begin{document}
\maketitle
\begin{abstract}
In today’s data-driven era, fully automated end-to-end data analytics, particularly insight discovery, is critical for discovering actionable insights that assist organizations in making effective decisions.
With the rapid advancement of large language models (LLMs), LLM-driven agents have emerged as a promising paradigm for automating data analysis and insight discovery. However, existing data insight agents remain limited in several key aspects, often failing to deliver satisfactory results due to: (1) insufficient utilization of domain knowledge, (2) shallow analytical depth, and (3) error-prone code generation during insight generation. To address these issues, we propose~\modelname, a novel multi-agent framework that incorporates three innovative features including external knowledge retrieval to enrich the analytical context, a multi-role debating mechanism to simulate diverse analytical perspectives and deepen analytical depth, and multi-path reasoning to improve
the accuracy of the generated code and insights. Extensive experiments on InsightBench demonstrate that~\modelname~consistently outperforms existing data insight agents across all difficulty levels, offering an effective solution for automated data insight discovery.
\end{abstract}
 
\section{Introduction}
Nowadays, data continues to grow in volume and complexity across domains~\cite{l2017machine,najafabadi2015deep}.
The importance of data and data analysis cannot be overstated. Data serves as the foundation for informed decision-making, providing the raw information needed to understand complex situations. Through data analysis, the raw information is transformed into interpretable patterns and trends~\cite{khan2024effective,abdul2024enhancing,ibeh2024data}.
Building on this, insight discovery~\cite{sahu2025insightbench} goes a step further to uncover actionable insights that can drive innovation, guide strategic decisions, and improve outcomes. It empowers organizations to optimize operations, tailor strategies, and remain competitive in dynamic, high-stakes contexts~\cite{steiner2022harnessing, colson2019ai, mcafee2012big}.

Traditional data analysis and insight discovery largely rely on manual efforts, which are extremely time-consuming~\cite{bean2022becoming,arora2015analytics}.
With the advancement of large language models (LLMs), LLM-driven agents have recently demonstrated the capability for automated insight discovery.
Existing data insight agents~\cite{perez2025llm,sahu2025insightbench,langchain2024pandas} commonly adopt a question-driven paradigm. These approaches typically first raise relevant analytical questions, then answer them using SQL or Python-based code execution, and finally summarize the results into coherent insights.

\begin{figure}[!t]
    \centering
    \includegraphics[width=\linewidth]{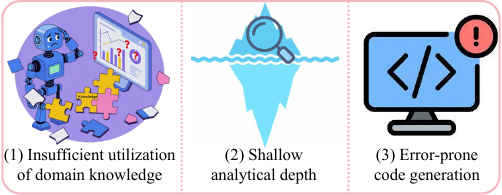}
    \caption{Illustration of three critical limitations in current data insight agents: (1) insufficient utilization of domain knowledge, (2) shallow analytical depth, and (3) error-prone code generation.}
    \label{fig:teaser}
\end{figure}

However, existing data insight agents are far from being as excellent as experienced data analysts in insight discovery.
As shown in Figure~\ref{fig:teaser}, there are at least three critical limitations in current approaches, as detailed in Appendix~\ref{app:error_analysis}.
(1) \textbf{Insufficient utilization of domain knowledge.} In real-world data analysis scenarios~\cite{sahu2025insightbench,majumder2024discoverybench,hu2024infiagent}, datasets often originate from diverse domains such as industry, healthcare, retail, etc., each characterized by domain-specific constraints and knowledge. Purely data-driven approaches~\cite{perez2025llm,sahu2025insightbench} or models relying solely on internal knowledge of LLMs often fail to capture critical domain constraints, external events, or expert heuristics that are essential for producing accurate and actionable insights.
(2) \textbf{Shallow analytical depth.} In question-driven approaches for insight discovery, the formulation of high-quality questions is as critical as answering them. Prior methods~\cite{perez2025llm,sahu2025insightbench,wu2024automated,10.1007/s00778-025-00912-0}~only rely on a single LLM to generate questions, which tends to result in shallow or overly generic questions, thereby limiting the depth and novelty of the generated insights.
(3) \textbf{Error-prone code generation.}
Code generation remains a challenging task for current LLMs~\cite{huynh2025large, hong2024next}. Relying on a single LLM to generate executable code~\cite{perez2025llm,sahu2025insightbench} often leads to errors, which can result in incorrect or misleading insights.


To alleviate above-mentioned issues, we propose~\modelname, a novel multi-agent framework that incorporates three innovative features: external knowledge retrieval, multi-role debating, and multi-path reasoning.
\modelname~is a modular multi-agent framework composed of four key modules that work in an iterative question-answering (QA) loop. \textbf{Dataset Description Module} provides the structured description of datasets. To address domain-specific challenges, \textbf{Retrieval-Augmented Knowledge Generation (RAKG) Module} dynamically retrieves and synthesizes external domain knowledge when internal knowledge of LLMs is insufficient. \textbf{Question Raising Module} formulates high-quality analytical questions through a divergent-convergent multi-role debating process, ensuring broad coverage and depth. These questions are passed to the \textbf{Insights Generation Module}, which translates questions into executable Python code through multi-path reasoning, interprets the outputs, and finally generates insights. Each core module adopts a multi-agent architecture to enhance specialization and collaboration. During iterative QA cycles, new questions are adaptively generated based on previous insight history, enabling deeper exploration. Finally, the generated insights are consolidated into a coherent summary. Extensive experiments on  InsightBench~\cite{sahu2025insightbench} demonstrate that~\modelname~consistently outperforms existing data agents across two metrics (i.e., insight-level and summary-level scores) and all difficulty levels.~\modelname~is also particularly well-suited for complex and high-difficulty tasks.

Our contributions can be summarized as follows:
\begin{itemize}
    \setlength{\itemsep}{0em}
    \item We propose~\modelname, a novel multi-agent framework composed of four key modules for automated insight discovery task.
    \item \modelname~incorporates three innovative features including external knowledge retrieval to enrich the analytical context, a multi-role debating mechanism to simulate diverse analytical perspectives and deepen analytical depth, and multi-path reasoning to improve the accuracy of the generated code and insights. 
    \item We conduct comprehensive experiments that demonstrate \modelname~consistently outperforms existing data insight agents, and is particularly well-suited for complex and high-difficulty tasks.
\end{itemize}


\begin{figure*}[th!]
    \centering
    \includegraphics[width=0.99\linewidth]{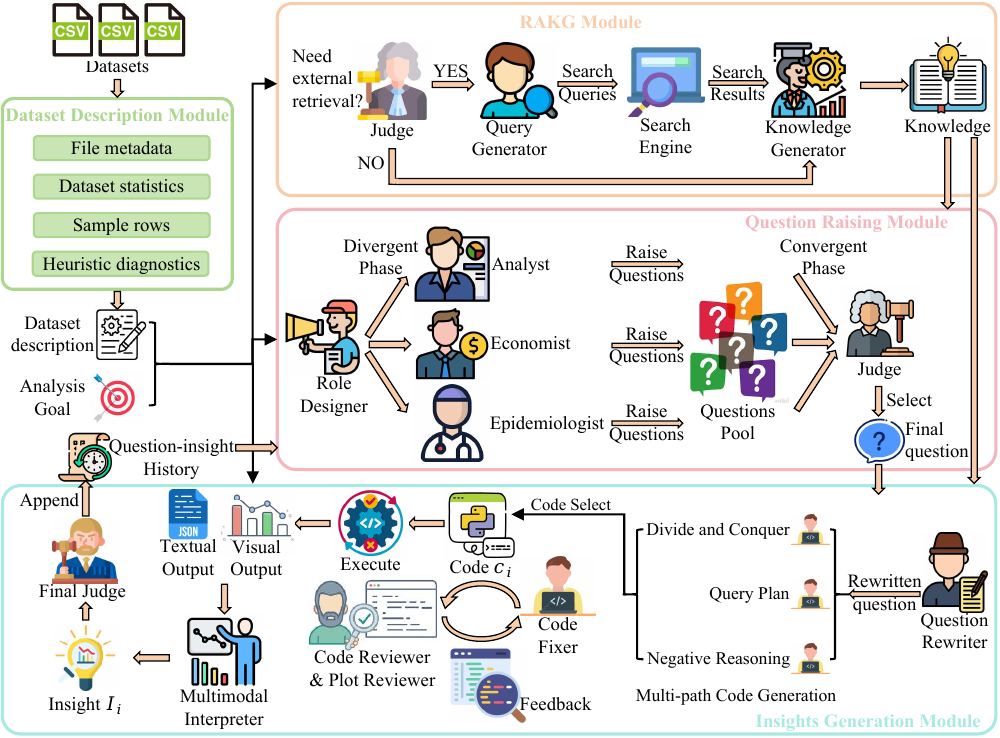}
    \caption{An illustration of {\modelname}, a multi-agent framework for insights discovery tasks. The framework consists of four key modules that work in an iterative QA loop. Dataset Description Module provides the structured description of datasets. RAKG Module retrieves and synthesizes external knowledge. Question Raising Module raises high-quality analytical questions through a divergent-convergent process. Finally, Insights Generation Module transforms questions into the final insights.}
    \label{fig:framework}
\end{figure*}

\section{Related Work}
\label{sec:bibtex}

\noindent \textbf{Data Analytics Agents.}
Recent years have seen rapid progress in the development of LLM-based agents for general-purpose data analysis. Systems such as Code Interpreter~\cite{openai2024code} and Pandas Agent~\cite{langchain2024pandas} allow users to analyze tabular data through natural language queries. \citeauthor{vacareanu2024words} demonstrates that LLMs can also perform basic statistical modeling such as regression, enhancing their applicability in quantitative tasks. Additional studies~\cite{cheng2023gpt,wang2023plan,wang2024executable,hong2024data} have proposed end-to-end LLM-based frameworks that combine goal understanding, code generation, or visualization to support analytical workflows.

\noindent \textbf{Insight Discovery Approaches.}
Insight discovery extends beyond basic data analysis by aiming to uncover meaningful, actionable insights from data. Early systems such as QuickInsights~\cite{ding2019quickinsights} and the template-based method by~\citeauthor{law2020characterizing} focus on extracting insights using predefined logic and simple heuristics. These approaches are often limited to clean datasets with well-labeled columns. With the advent of LLMs, newer systems such as InsightPilot~\cite{ma-etal-2023-insightpilot}, OpenAI Data Analysis~\cite{openai2024dataanalysis}, Langchain’s Pandas Agent~\cite{langchain2024pandas}, and HLI~\cite{perez2025llm} have emerged, which can generate code to present descriptive insights. These methods often specialize in narrow and single-step insight discovery tasks based on very concrete user instructions. AgentPoirot~\cite{sahu2025insightbench} extends this line of work toward multi-step insight generation, which achieves state-of-the-art performance on the InsightBench benchmark~\cite{sahu2025insightbench}. However, there are some persistent limitations in existing systems, such as insufficient utilization of domain knowledge, limited analytical depth, and error-prone code generation. We introduce~\modelname, a multi-agent framework featuring external knowledge retrieval, multi-role debating, and multi-path reasoning, pushing beyond the boundaries of current LLM-based systems.

\section{The {\modelname} Framework}

\subsection{Framework Overview}
As illustrated in Figure~\ref{fig:framework}, our framework adopts a question-driven paradigm to systematically generate insights from datasets. The framework consists of four key modules that work in an iterative QA loop with $N_{\text{iter}}$ iterations. \textbf{Dataset Description Module} provides the structured description of datasets that serves as foundational context for downstream modules. \textbf{Retrieval-Augmented Knowledge Generation (RAKG) Module} dynamically retrieves and synthesizes external knowledge when internal knowledge of LLMs is insufficient. \textbf{Question Raising Module} formulates high-quality analytical questions through a divergent-convergent process, ensuring broad coverage and depth. \textbf{Insights Generation Module} transforms questions into executable Python code, rigorously validates the code, runs the code in the sandbox, interprets the output of the code, and finally produces insights. Each core module (except Dataset Description) employs a multi-agent architecture to enhance specialization. During iterative QA cycles, subsequent questions are dynamically raised based on prior question-insight history. Finally, all insights are consolidated into a coherent summary.

\subsection{Dataset Description Module}
To support rapid understanding of the datasets, we design a unified dataset description module, which can automatically extract essential metadata and detect potential issues in the datasets. Given the raw dataset $\mathbb{D}$, the module first collects basic file metadata including the filename, file size, and file type. Then, it computes comprehensive dataset statistics, including dimensionality (number of rows and columns), column-wise data types, and per-column missing value counts. If the dataset contains numeric columns, it further computes standard descriptive statistics such as mean, standard deviation, min/max, and quantiles. A representative sample of the first few rows is also extracted. The module also includes lightweight heuristic diagnostics, such as flagging missing values or duplicated rows. All the aforementioned information is organized into a structured JSON format, serving as the comprehensive dataset description $D$. This abstraction simplifies dataset onboarding, facilitates downstream understanding of the dataset, and enables consistent preprocessing pipelines across heterogeneous data sources.

\subsection{Retrieval Augmented Knowledge Generation (RAKG) Module}
To address domain-specific challenges, we equip our framework with an \textit{on-demand} external retrieval mechanism to detect, retrieve, and synthesize external domain knowledge to support the whole analysis pipeline.
Our RAKG module consists of the following four main stages.

\noindent\textbf{Search Necessity Judgment.}
Given dataset description $D$ and analysis goal $G$, a \textit{judge} agent determines whether the analysis can be completed using only internal knowledge and the provided datasets.
\begin{equation}
\mathrm{Judge}(D, G) \to \{\text{yes}, \text{no}\},
\end{equation}

If the \textit{judge} determines external retrieval is unnecessary, the module defaults to a \textit{vanilla knowledge generator}, leveraging only the internal knowledge of LLMs. Otherwise, it proceeds to the next stage.

\noindent\textbf{Search Query Generation.}
For cases where external retrieval is needed, a \textit{data-aware query generator} formulates $N_q$ high-quality, Google-ready search queries $\mathcal{Q}$. These queries are crafted to target vertical knowledge gaps identified in the judgment stage, aiming for diversity and minimal redundancy.
\begin{equation}
\mathcal{Q} = \mathrm{QueryGenerator}(D, G),
\end{equation}

\noindent\textbf{Search Execution.}
The module then performs real-time web retrieval using Google Search~\footnote{Serper API: https://serper.dev/ .}. To ensure reproducibility, we impose a maximum date constraint on the search.
\begin{equation}
\mathcal{R} = \bigcup_{q_s \in \mathcal{Q}} \mathrm{GoogleSearch}(q_s),
\end{equation}

\noindent\textbf{Knowledge Generation.}
A specialized \textit{knowledge generator} processes the search results $\mathcal{R}$ and produces structured, domain-relevant knowledge items $K$. These knowledge items are then made available to downstream modules, such as Question Raising and Question Rewriting.
\begin{equation}
K = \mathrm{KnowledgeGenerator}(\mathcal{R}, D, G),
\end{equation}
\subsection{Question Raising Module}
Inspired by cognitive theories of {divergent–convergent thinking}~\cite{lu2024llm}, we propose a structured Question Raising Module that adopts a multi-agent, role-driven framework to simulate diverse analytical perspectives and deepen analytical depth. The design of the module embraces a divergent-convergent paradigm: agents first explore broadly from distinct perspectives (divergent phase) and then select the most promising questions (convergent phase). The following are the three key stages in the module.

\noindent \textbf{Data-Derived Role Generation.}
Given dataset description $D$, analysis goal $G$ and knowledge $K$, we prompt a system-level \textit{role designer} agent to generate $N_R$ diverse and well-specified analytical roles. Each role is defined by attributes such as its background (e.g., behavioral analyst, anomaly detector), domain focus, personality traits (e.g., skeptical, risk-seeking), and analytical capabilities. The \textit{role designer} ensures that the roles align with the datasets and analysis goal, and aims to maximize coverage over the question space.
\begin{equation}
\{\mathrm{Role}_1, \dots, \mathrm{Role}_{N_R}\} = \mathrm{RoleDesigner}(D, G, K),
\end{equation}

\noindent\textbf{Multi-Role Question Raising.}
Each generated role then independently explores the data and poses a set of questions aligned with its unique perspective. Roles may specialize in temporal dynamics, user behavior, value distribution, or rare event detection, among others. This results in a pool of questions $\mathbb{Q}_j$ (in the $j$-th iteration) that span a wide variety of analytical angles, ranging from descriptive to causal, comparative to behavioral.
\begin{equation}
\mathbb{Q}_j = \bigcup_{1 \leq i \leq N_R} \mathrm{Role_i}(D, G, K, H),
\end{equation}
where $H$ represents the history of previous question-insight pairs.

\noindent \textbf{Question Convergence.}
The convergent phase is introduced to select the most promising questions in the question pool $\mathbb{Q}_j$. A global \textit{judge} agent selects a subset of high-quality questions $\mathbb{Q}^*_j$ based on the following criteria: (1) Potential to yield non-trivial or surprising insights; (2) Alignment with datasets and analysis goal; (3) Diversity across question types and dimensions; (4) Complementarity with already answered questions. Each selected question is annotated with its source role and a justification for selection, making the reasoning process transparent and interpretable.
\begin{equation}
\mathbb{Q}^*_j = \mathrm{Judge}(\mathbb{Q}_j, D, G, K, H) \subseteq \mathbb{Q}_j,
\end{equation}
\subsection{Insights Generation Module}
Once a question $q \in \mathbb{Q}^*_j$ is proposed by the Question Raising Module, the Insights Generation Module takes over to generate the corresponding insight.
Naively answering natural language questions with code generation often leads to failures due to ambiguity~\cite{zhou2023learned}, schema mismatch~\cite{wang2025linkalign}, or unvalidated code logic~\cite{cen2025sqlfixagent}.
To ensure the quality of the generated insight, we decompose the insight generation into a structured pipeline that includes Question Clarification, Multi-path Code Generation and Refinement, Multimodal Insight Interpretation, and Final Decision.

\noindent \textbf{Question Clarification.}
To reduce ambiguity in questions and enhance code generation accuracy, we first employ a \textit{schema-aware question rewriter}. The \textit{rewriter} identifies missing table and column references, unclear aggregation goals, temporal dimensions, or ambiguous variable names, and reformulates the question into a fully grounded, self-contained analytical question $q^*$.
\begin{equation}
q^* = \mathrm{QuestionRewriter}(q, D, G, K),
\end{equation}

\noindent \textbf{Multi-path Code Generation.}
To increase the likelihood of generating correct code, we adopt a multi-path code generation strategy.
Inspired by the success of Chain-of-Thought (CoT) prompting~\cite{wei2022chain}~in improving the reasoning capabilities of large language models (LLMs), we adopt three complementary CoT reasoning strategies tailored for code generation: \textbf{Divide-and-Conquer}, \textbf{Query Plan}, and \textbf{Negative Reasoning}.

The Divide-and-Conquer Reasoning decomposes a complex problem into smaller sub-problems, solves each sub-problem independently, and then combines the individual solutions to obtain the final code $c_{DaC}$.

The Query Plan Reasoning first produces a query execution plan in natural language (detailing filters, joins, aggregations, and intermediate outputs) before translating the plan into executable code $c_{QP}$.

The Negative Reasoning anticipates plausible mistakes (e.g., double counting, incorrect joins, handling of NULLs), explains how to avoid them, and only then generates code $c_{NR}$ designed to mitigate those risks.

By leveraging these three reasoning paths, we generate a diverse pool of candidate code.
Among the candidates, a \textit{code selector} is employed to evaluate and select the most appropriate one as the initial code $c_0$ for the subsequent refinement.
\begin{equation}
c_0 = \mathrm{CodeSelector}(q^*, D, G, c_{DaC},c_{QP},c_{NR}),\\
\end{equation}

\noindent \textbf{Code Refinement.} To ensure correctness, the initial code $c_0$ is passed through a \textit{code reviewer}. The \textit{code reviewer} analyzes the code along four dimensions involves requirement alignment, schema compliance, operational risk, and data integrity.
\begin{equation}
r_{c_i} = \mathrm{CodeReviewer}(q^*, D, G, c_i),
\end{equation}

Furthermore, the generated code $c_0$ is executed to produce a plot that visualizes the results of data analysis. However, the initial plot $p_0$ is often of low quality and difficult to interpret, due to issues such as overlapping text or poor layout. To improve the readability and interpretability of these plots, both for human and downstream interpreter model, we further introduce a \textit{plot reviewer} that evaluates the generated plots.
\begin{equation}
r_{p_i} = \mathrm{PlotReviewer}(q^*, D, G, p_i),
\end{equation}

If issues are found (either by the \textit{code reviewer} or the \textit{plot reviewer}), the \textit{code fixer} takes the original code $c_0$ and the reviewers’ feedback $r_0=\{r_{c_0}, r_{p_0}\}$ to produce a revised version $c_1$. This process may iterate several times, particularly for complex or error-prone questions until no more mistakes are found or a predefined maximum number of iterations ($N_{\mathrm{fix}}$) is reached.
\begin{equation}
\begin{aligned}
c_i &= \mathrm{CodeFixer}(q^*, D, G, c_{i-1}, r_{i-1}),\\
 &\quad \forall\, 1 \leq i \leq N_{\mathrm{fix}}, \mathrm{FAIL} \in r_{i-1},
\end{aligned}
\end{equation}

\noindent \textbf{Multimodal Insight Interpretation and Final Decision.}
Once a code version $c_i$ is obtained, it is executed in the sandbox. Then, the \textit{interpreter} agent parses the generated visual/textual artifacts and generates the insight $I_i$.
Previous methods~\cite{perez2025llm,sahu2025insightbench}~rely solely on textual information to generate insights. However, textual information alone is often insufficient for accurately explaining analytical results. In contrast, since we explicitly review and refine the generated plot in the preceding step, the resulting visualization is more interpretable. To this end, we are the first to use Multimodal Large Language Models (MLLMs) as the interpreter that jointly leverages textual and visual information when generating insights.

To avoid the possible introduction of new errors during the code refinement process, all intermediate versions of the code ${c_i}$ and corresponding insights ${I_i}$ are stored and passed to a \textit{final judge}, which selects the most complete, valid, and interpretable insight $I$ according to the history of code refinement. Finally, the $q-I$ pair is added to the history $H$.
\begin{equation}
\begin{aligned}
o_i &= \mathrm{SandboxExecute}(c_i), \\
I_i &= \mathrm{Interpreter}(q,D,o_i), \\
I &= \mathrm{FinalJudge}(q,D,G, \{(c_i,I_i)\}), \\
H &= H \cup \{(q, I)\}, \\
\end{aligned}
\end{equation}

\begin{table*}[t!]
\centering
\setlength{\tabcolsep}{6pt} 
\renewcommand{\arraystretch}{1.0} 
\begin{tabular}{l|cccc|cccc}
\toprule
\multirow{2}{*}{\textbf{Model}} & \multicolumn{4}{c|}{\textbf{Insight-level Scores (G-Eval)}} & \multicolumn{4}{c}{\textbf{Summary-level Scores (G-Eval)}} \\
\cline{2-5}
\cline{6-9}
 & Easy & Medium & Hard & Avg & Easy & Medium & Hard & Avg \\
\midrule
\multicolumn{9}{l}{\textit{LLM-only}} \\
GPT-4o only & 0.3157 & 0.2572 & 0.2702 & 0.2789 & 0.3622 & 0.3141 & 0.2923 & 0.3215 \\
GPT-4o domain & 0.3195 & 0.2608 & 0.2663 & 0.2802 & 0.3649 & 0.3144 & 0.3164 & 0.3302 \\
\midrule
\multicolumn{9}{l}{\textit{Single-agent Models}} \\
CodeGen & 0.3268  & 0.2766 & 0.2563 & 0.2852 & 0.3424 & 0.3023  & 0.2773 & 0.3063 \\
ReAct & 0.3355 & 0.2953 & 0.2672 & 0.2984 & 0.3618 & 0.3239 & 0.2714 & 0.3185 \\
\midrule
\multicolumn{9}{l}{\textit{Multi-agent Models}} \\
Data-to-Dashboard & 0.2191 & 0.2148 & 0.2368 & 0.2231 & 0.2463 & 0.2514 & 0.2615 &  0.2531 \\
Pandas Agent & 0.3479 & 0.2796 & \underline{0.3180} & 0.3124  & 0.3589 & 0.3298 & 0.2998 & 0.3289 \\
AgentPoirot & \underline{0.3768} & \underline{0.3044} & 0.3117 & \underline{0.3284} & \underline{0.3819} & \underline{0.3595} & \underline{0.3292} & \underline{0.3565} \\
\midrule
\multirow{2}{*}{\modelname~(Ours)} & \textbf{0.4063} & \textbf{0.3211} & \textbf{0.3407} & \textbf{0.3530} & \textbf{0.4448} & \textbf{0.4047} & \textbf{0.3710} & \textbf{0.4059} \\
&\textbf{+7.8\%} & \textbf{+5.5\%} & \textbf{+9.3\%} & \textbf{+7.5\%} &\textbf{+16.5\%}&\textbf{+12.6\%}&\textbf{+12.7\%}&\textbf{+13.9\%}\\
\bottomrule
\end{tabular}
\caption{Performance of different models on the InsightBench. The best and runner-up are in \textbf{bold} and \underline{underlined}. Our framework consistently outperforms the best baseline across both metrics and all three difficulty levels. Improvements are calculated between the best to the runner-up.}
\label{tab:main_results}
\end{table*}

\section{Experiments}
\subsection{Experiment Setup}
\noindent\textbf{Dataset.}\quad
We utilize the InsightBench~\cite{sahu2025insightbench}, a widely used benchmark for evaluating insight discovery in data analytics. It consists of 100 tabular datasets representing diverse business use cases, spanning three difficulty levels: easy, medium, and hard. Unlike other datasets that focus on more specific QA style data analysis tasks~\cite{hu2024infiagent,majumder2024discoverybench}, InsightBench evaluates agents on their ability to perform end-to-end data analytics, encompassing question formulation, answer interpretation, and insight generation.

\noindent\textbf{Baselines.}\quad
We compare our framework against baselines of three categories: 1) \textbf{LLM-only.} \textbf{GPT-4o only} is a direct prompting baseline where the dataset description $D$ and analysis goal $G$ are provided to GPT-4o without intermediate reasoning or tool use. The model is asked to directly generate the final set of insights. \textbf{GPT-4o domain} is a domain-aware GPT-4o baseline implemented by~\newcite{zhang2025data} that first infers the dataset’s domain, generates relevant domain knowledge, and then leverages this knowledge to produce the final insights. 2) \textbf{Single-agent Models.} \textbf{CodeGen}~\cite{majumder2024discoverybench} generates the entire code at one go to solve the task, where a demonstration of a solution code is provided in the context. Based on the execution result, it generates the insights and summarizes the workflow. \textbf{ReAct}~\cite{yao2023react} solves the task by generating thought and subsequent codes in a multi-turn fashion. 3) \textbf{Multi-agent Models.} \textbf{Data-to-Dashboard}~\cite{zhang2025data} is a modular multi-agent LLM system that automates end-to-end dashboard generation from tabular data. It integrates domain-aware reasoning with iterative self-reflection to produce insightful visualizations. We take its intermediate insight outputs as the results. \textbf{Pandas Agent}~\cite{langchain2024pandas} is a LangChain-based data science agent optimized for question answering. Given a data frame and a question, it generates and executes Python code to produce answers. \textbf{AgentPoirot}~\cite{sahu2025insightbench} is the baseline data analysis agent in InsightBench. It adopts a question-driven paradigm, which first generates root questions and then produces follow-up questions based on each root question.


\noindent\textbf{Metrics.}\quad
We follow the InsightBench~\cite{sahu2025insightbench} evaluation protocol, which computes performance at two levels: (1) \textit{Summary-level}, measuring the LLaMA-3-Eval score between the generated and ground-truth summaries; and (2) \textit{Insight-level}, matching each ground-truth insight with the most similar prediction and averaging their scores. In our experiments, we adopt G-Eval~\cite{liu2023g} instead of LLaMA-3-Eval for better alignment with human judgment and discard ROUGE-1 due to its well-known limitations in capturing semantic similarity~\cite{chen2024rethinking}. We report the average of the summary-level and insight-level scores across datasets of different difficulty levels.

\noindent\textbf{Implementation Details.}\quad
To ensure a fair comparison across models, we use GPT-4o~\footnote{Version gpt-4o-2024-08-06} as the base model and the temperature is fixed to $0.0$ for all baselines. We set the hyperparameters as follows: $N_q=3$ (number of search queries in RAKG Module), $N_R=3$ (number of diverse analytical roles in Question Raising Module), $N_{\text{fix}}=5$ (maximum number of code-fix iterations), and $N_{\text{iter}}=6$ (number of Q-A loop iterations, consistent with the baseline AgentPoirot for fair comparison). The prompts and implementation details for each agent are provided in the Appendix~\ref{app:prompt}.

\subsection{Main Results}

We compare our framework with three types of baselines and present the results in Table~\ref{tab:main_results}. The key findings from the table are as follows: (1) \textbf{Consistent performance improvement with our framework.} Our framework consistently outperforms the best baseline across both metrics and all three difficulty levels. On Insight-level Scores,~\modelname~achieves an average improvement of +7.5\%, while on Summary-level Scores the gain is even larger at +13.9\%. (2) \textbf{Larger improvements on harder datasets.} The performance improvement is more pronounced on more challenging tasks. On Insight-level Scores, our framework improves over the best baseline by +7.8\%, +5.5\%, and +9.3\% for Easy, Medium, and Hard datasets, respectively. This demonstrates that our framework is particularly well-suited for complex, high-difficulty tasks, while still providing improvements on simpler datasets, indicating strong generality. (3) \textbf{More substantial gains on Summary-level Scores.} The improvement of~\modelname~on Summary-level Scores is notably larger than that on Insight-level Scores, despite no explicit optimization for summary generation. This suggests that producing more accurate insights enables the model to focus on the correct information during summarization, thereby improving summary quality. (4) \textbf{Advantages of multi-agent architectures.} Across the three baseline categories, multi-agent models achieve the best performance in insight discovery (except Data-to-Dashboard, which does not fully address the insight discovery task), followed by single-agent models, with LLM-only approaches performing the worst. This aligns with intuition and highlights the inherent strengths of multi-agent designs. Our carefully designed~\modelname~further amplifies these advantages, achieving the highest performance among all structures. (5) \textbf{Effectiveness of domain-aware prompting.} \textit{GPT-4o domain} slightly outperforms \textit{GPT-4o only}, indicating that even a simple domain-aware design can enhance performance. This underscores the importance of domain-specific knowledge for this task and further validates the rationale behind the RAKG module in our framework.
\subsection{Quality of Plots}

\begin{figure}[t!]
    \centering
    \includegraphics[width=0.95\linewidth]{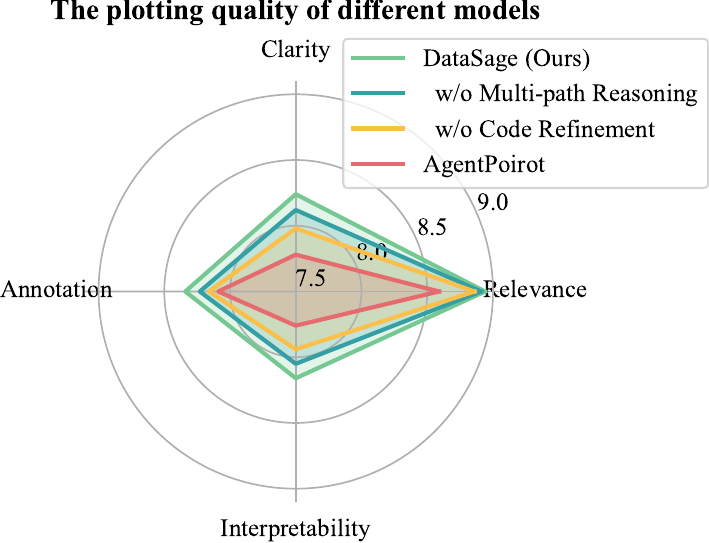}
    \caption{Comparison of the plot quality across different models. \modelname~can generate noticeably higher-quality plots than the baseline AgentPoirot, attributing to our design of the Insights Generation Module.}
    \label{fig:plot_quality}
    \vspace{-5pt}
\end{figure}

Beyond producing superior insights, \modelname~can also generate plots with higher quality. To demonstrate this, we compare~\modelname~against the best baseline AgentPoirot, as well as two ablated variants that remove either the \textit{Code Refinement} or \textit{Multi-path Reasoning} component. To enable large-scale and consistent evaluation, we employ GPT-4o as a judge to score the generated plot for each raised question along four dimensions: (1) \textbf{Relevance}: whether the plot accurately addresses the raised question; (2) \textbf{Clarity}: whether the visualization is clean, legible, and free of unnecessary clutter; (3) \textbf{Annotation}: whether axis labels, titles, legends, and color usage are correct, informative, distinct, and accessible; and (4) \textbf{Interpretability}: whether a viewer can readily identify and articulate the key takeaway of the plot. Each criterion is rated on a 0–10 scale. If the model fails to generate a plot, it is assigned a score of 0.

As shown in Figure~\ref{fig:plot_quality}, \modelname~consistently outperforms the baseline by a substantial margin across all four dimensions. The ablation results confirm that both \textit{Code Refinement} and \textit{Multi-path Reasoning} contribute to this improvement, with \textit{Code Refinement} having a more pronounced effect. It demonstrates that the design of our Insights Generation Module can not only enhance the correctness of generated code but also improve the quality of the resulting plots. A more detailed case study can be found in Appendix~\ref{app:case_study_plot}. \modelname~can produce higher-quality plots that not only enable downstream \textit{interpreter} agent to generate insights more accurately but also facilitate human comprehension. It enhances both the utility and the trustworthiness of our framework, empowering users to make informed decisions more effectively.

\subsection{Ablation Study}
\begin{table}[t!]
\centering
\setlength{\tabcolsep}{3.5pt} 
\begin{tabular}{lcc}
\toprule
\multirow{2}{*}{\textbf{Model}} & \multicolumn{2}{c}{\textbf{G-Eval}} \\
\cline{2-3}
& \textbf{Insight-level} & \textbf{Summary-level} \\
\midrule
\modelname & \textbf{0.3530} & \textbf{0.4059} \\
\quad w/o QR & 0.3475 & 0.4019 \\
\quad w/o MR & 0.3417 & 0.3948 \\
\quad w/o RAKG & 0.3316 & 0.3982 \\
\bottomrule
\end{tabular}
\caption{Ablation study on the three key components of our framework: Retrieval-Augmented Knowledge Generation (RAKG), Question Raising (QR), and Multi-path Reasoning (MR). The best results are in \textbf{bold}. Removing any component degrades performance, with RAKG contributing most to the overall performance.}
\label{tab:ablation_study_overall}
\end{table}
To validate the effectiveness of each proposed component in~\modelname, we conduct an ablation study by removing one of the three key modules, Retrieval-Augmented Knowledge Generation (RAKG), Question Raising (QR), and Multi-path Reasoning (MR), while keeping the rest of the framework unchanged.

As shown in Table~\ref{tab:ablation_study_overall}, ablating any component results in a performance drop over all the metrics. This indicates that all the proposed components contribute to the superior performance of~\modelname, validating the rationality of our framework design. Removing RAKG results in the most substantial degradation in overall performance. This suggests that grounding the analysis with external knowledge is crucial for generating deep and accurate insights. Overall, the ablation study confirms that all three components are indispensable and complementary. Their combined effect leads to the highest performance. More detailed ablation analysis and hyperparameter study is provided in Appendix~\ref{app:experimental_results}.
Additionally, the case study is presented in Appendix~\ref{app:case_study}.



\section{Conclusion}
In this paper, we propose~\modelname, which demonstrates the potential of a multi-agent framework to significantly enhance data insight discovery by addressing key limitations of existing data agent systems through external knowledge retrieval, multi-role debating, and multi-path reasoning. Our approach not only improves the accuracy and reliability of insights but also offers a scalable and flexible solution for automated data analysis. Future work will focus on further optimizing the framework, designing other agent architectures, and exploring additional applications in diverse domains.

\section*{Limitations}
While our proposed framework achieves state-of-the-art performance on the insight discovery task, several limitations remain.

First, despite outperforming existing baselines, there is still a gap between our framework and experienced data analysts in terms of analytical depth, contextual understanding, and the ability to draw nuanced insights.    We view this as an important direction for future work and plan to further enhance the reasoning capabilities and domain alignment of the framework.

Second, \modelname~is primarily designed and evaluated for insight discovery tasks.    Although this task is representative of many real-world analytical scenarios, it only covers a subset of the broader data analysis landscape.    In future work, we aim to extend our framework to support a wider range of analytical tasks, exploring additional applications across diverse domains and data modalities.

Third, \modelname~is particularly well-suited for complex and high-difficulty analytical problems, where multi-path reasoning and external knowledge integration are crucial.    However, for relatively simple tasks, some of the modules may be redundant and could introduce unnecessary computational or interpretive overhead.    As a result, we plan to investigate adaptive mechanisms that can dynamically tailor the analysis pipeline based on task difficulty, thereby improving overall efficiency and usability.
\section*{Ethical Statement}
Our work aims to improve the efficiency and scalability of data analysis by automating the generation of analytical insights from structured datasets. This can be particularly valuable in settings where manual analysis is costly or infeasible, enabling organizations to optimize operations and tailor strategies more effectively.

However, we acknowledge that the insights generated by our framework~\modelname~are produced automatically and may contain inaccuracies, misinterpretations, or incomplete reasoning due to model limitations or data quality issues. As such, the generated insights should be treated as preliminary references rather than definitive conclusions. Users must exercise critical judgment and, when necessary, seek human expert verification before acting on any insights produced by~\modelname, especially in high-stakes domains such as finance, healthcare, or policy-making.

We emphasize that~\modelname~is intended to serve as an assistive tool to augment human analytical capabilities, not to replace domain experts or rigorous manual analysis. Responsible use requires transparency about the framework’s limitations and active user oversight to avoid unintended consequences or over-reliance on automated outputs.
By clearly framing~\modelname~as a decision support tool and not a decision maker, we aim to encourage responsible deployment and maximize its potential benefits while minimizing misuse.


\bibliography{custom}

@inproceedings{
sahu2025insightbench,
title={InsightBench: Evaluating Business Analytics Agents Through Multi-Step Insight Generation},
author={Gaurav Sahu and Abhay Puri and Juan A. Rodriguez and Amirhossein Abaskohi and Mohammad Chegini and Alexandre Drouin and Perouz Taslakian and Valentina Zantedeschi and Alexandre Lacoste and David Vazquez and Nicolas Chapados and Christopher Pal and Sai Rajeswar and Issam H. Laradji},
booktitle={The Thirteenth International Conference on Learning Representations},
year={2025},
url={https://openreview.net/forum?id=ZGqd0cbBvm}
}

@inproceedings{perez2025llm,
  title={An LLM-Based Approach for Insight Generation in Data Analysis},
  author={P{\'e}rez, Alberto S{\'a}nchez and Boukhary, Alaa and Papotti, Paolo and Lozano, Luis Castej{\'o}n and Elwood, Adam},
  booktitle={Proceedings of the 2025 Conference of the Nations of the Americas Chapter of the Association for Computational Linguistics: Human Language Technologies (Volume 1: Long Papers)},
  pages={562--582},
  year={2025}
}

@article{wu2024automated,
  title={Automated data visualization from natural language via large language models: An exploratory study},
  author={Wu, Yang and Wan, Yao and Zhang, Hongyu and Sui, Yulei and Wei, Wucai and Zhao, Wei and Xu, Guandong and Jin, Hai},
  journal={Proceedings of the ACM on Management of Data},
  volume={2},
  number={3},
  pages={1--28},
  year={2024},
  publisher={ACM New York, NY, USA}
}

@article{10.1007/s00778-025-00912-0,
author = {Li, Shuaimin and Chen, Xuanang and Song, Yuanfeng and Song, Yunze and Zhang, Chen Jason and Hao, Fei and Chen, Lei},
title = {Prompt4Vis: prompting large language models with example mining for tabular data visualization},
year = {2025},
issue_date = {Jul 2025},
publisher = {Springer-Verlag},
address = {Berlin, Heidelberg},
volume = {34},
number = {4},
issn = {1066-8888},
url = {https://doi.org/10.1007/s00778-025-00912-0},
doi = {10.1007/s00778-025-00912-0},
journal = {The VLDB Journal},
month = may,
numpages = {26}
}

@inproceedings{lu2024llm,
  title={LLM Discussion: Enhancing the Creativity of Large Language Models via Discussion Framework and Role-Play},
  author={Lu, Li-Chun and Chen, Shou-Jen and Pai, Tsung-Min and Yu, Chan-Hung and Lee, Hung-yi and Sun, Shao-Hua},
  booktitle={First Conference on Language Modeling},
    year={2024}
}

@article{wei2022chain,
  title={Chain-of-thought prompting elicits reasoning in large language models},
  author={Wei, Jason and Wang, Xuezhi and Schuurmans, Dale and Bosma, Maarten and Xia, Fei and Chi, Ed and Le, Quoc V and Zhou, Denny and others},
  journal={Advances in neural information processing systems},
  volume={35},
  pages={24824--24837},
  year={2022}
}

@article{zhang2025data,
  title={Data-to-Dashboard: Multi-Agent LLM Framework for Insightful Visualization in Enterprise Analytics},
  author={Zhang, Ran and Elhamod, Mohannad},
  journal={arXiv preprint arXiv:2505.23695},
  year={2025}
}

@inproceedings{hu2024infiagent,
  title={InfiAgent-DABench: Evaluating Agents on Data Analysis Tasks},
  author={Hu, Xueyu and Zhao, Ziyu and Wei, Shuang and Chai, Ziwei and Ma, Qianli and Wang, Guoyin and Wang, Xuwu and Su, Jing and Xu, Jingjing and Zhu, Ming and others},
  booktitle={International Conference on Machine Learning},
  pages={19544--19572},
  year={2024},
  organization={PMLR}
}

@inproceedings{majumder2024discoverybench,
  title={Discoverybench: Towards data-driven discovery with large language models},
  author={Majumder, Bodhisattwa Prasad and Surana, Harshit and Agarwal, Dhruv and Mishra, Bhavana Dalvi and Meena, Abhijeetsingh and Prakhar, Aryan and Vora, Tirth and Khot, Tushar and Sabharwal, Ashish and Clark, Peter},
  booktitle={International Conference on Learning Representations (ICLR)},
  year={2025}
}

@inproceedings{yao2023react,
  title={React: Synergizing reasoning and acting in language models},
  author={Yao, Shunyu and Zhao, Jeffrey and Yu, Dian and Du, Nan and Shafran, Izhak and Narasimhan, Karthik and Cao, Yuan},
  booktitle={International Conference on Learning Representations (ICLR)},
  year={2023}
}

@misc{langchain2024pandas,
  author       = {{LangChain}},
  title        = {Pandas DataFrame},
  year         = {2024},
  howpublished = {\url{https://python.langchain.com/v0.2/docs/integrations/toolkits/pandas/}},
  note         = {Accessed: 2025-08-04}
}

@inproceedings{liu2023g,
    title = "{G}-Eval: {NLG} Evaluation using Gpt-4 with Better Human Alignment",
    author = "Liu, Yang  and
      Iter, Dan  and
      Xu, Yichong  and
      Wang, Shuohang  and
      Xu, Ruochen  and
      Zhu, Chenguang",
    editor = "Bouamor, Houda  and
      Pino, Juan  and
      Bali, Kalika",
    booktitle = "Proceedings of the 2023 Conference on Empirical Methods in Natural Language Processing",
    month = dec,
    year = "2023",
    address = "Singapore",
    publisher = "Association for Computational Linguistics",
    url = "https://aclanthology.org/2023.emnlp-main.153/",
    doi = "10.18653/v1/2023.emnlp-main.153",
    pages = "2511--2522"
}

@article{chen2024rethinking,
  title={Rethinking scientific summarization evaluation: grounding explainable metrics on facet-aware benchmark},
  author={Chen, Xiuying and Wang, Tairan and Zhu, Qingqing and Guo, Taicheng and Gao, Shen and Lu, Zhiyong and Gao, Xin and Zhang, Xiangliang},
  journal={ArXiv},
  pages={arXiv--2402},
  year={2024}
}

@article{mcafee2012big,
  title={Big data: the management revolution},
  author={McAfee, Andrew and Brynjolfsson, Erik and Davenport, Thomas H and Patil, DJ and Barton, Dominic},
  journal={Harvard business review},
  volume={90},
  number={10},
  pages={60--68},
  year={2012},
  publisher={Cambridge}
}

@article{colson2019ai,
  title={What AI-driven decision making looks like},
  author={Colson, Eric},
  journal={Harvard Business Review},
  volume={8},
  pages={2--8},
  year={2019}
}

@article{bean2022becoming,
  title={Why becoming a data-driven organization is so hard},
  author={Bean, Randy},
  journal={Harvard Business Review},
  year={2022}
}

@article{huynh2025large,
  title={Large Language Models for Code Generation: A Comprehensive Survey of Challenges, Techniques, Evaluation, and Applications},
  author={Huynh, Nam and Lin, Beiyu},
  journal={arXiv preprint arXiv:2503.01245},
  year={2025}
}

@article{hong2024next,
  title={Next-generation database interfaces: A survey of llm-based text-to-sql},
  author={Hong, Zijin and Yuan, Zheng and Zhang, Qinggang and Chen, Hao and Dong, Junnan and Huang, Feiran and Huang, Xiao},
  journal={arXiv preprint arXiv:2406.08426},
  year={2024}
}

@article{steiner2022harnessing,
  title={Harnessing data to make betterinformed decisions},
  author={Steiner, Stefan},
  journal={Scientia},
  year={2022}
}

@inproceedings{arora2015analytics,
  title={Analytics: Key to go from generating big data to deriving business value},
  author={Arora, Deepali and Malik, Piyush},
  booktitle={2015 IEEE first international conference on big data computing service and applications},
  pages={446--452},
  year={2015},
  organization={IEEE}
}

@article{l2017machine,
  title={Machine learning with big data: Challenges and approaches},
  author={L’heureux, Alexandra and Grolinger, Katarina and Elyamany, Hany F and Capretz, Miriam AM},
  journal={Ieee Access},
  volume={5},
  pages={7776--7797},
  year={2017},
  publisher={IEEE}
}

@article{najafabadi2015deep,
  title={Deep learning applications and challenges in big data analytics},
  author={Najafabadi, Maryam M and Villanustre, Flavio and Khoshgoftaar, Taghi M and Seliya, Naeem and Wald, Randall and Muharemagic, Edin},
  journal={Journal of big data},
  volume={2},
  number={1},
  pages={1},
  year={2015},
  publisher={Springer}
}

@article{khan2024effective,
  title={Effective decision making using data analytics},
  author={Khan, Attia},
  journal={Indian Scientific Journal Of Research In Engineering And Management},
  volume={8},
  number={04},
  pages={1--5},
  year={2024}
}

@article{abdul2024enhancing,
  title={Enhancing business performance: The role of data-driven analytics in strategic decision-making},
  author={Abdul-Azeez, Oluwatosin and Ihechere, Alexsandra Ogadimma and Idemudia, Courage},
  journal={International Journal of Management \& Entrepreneurship Research},
  volume={6},
  number={7},
  pages={2066--2081},
  year={2024}
}

@article{ibeh2024data,
  title={Data analytics in healthcare: A review of patient-centric approaches and healthcare delivery},
  author={Ibeh, Chidera Victoria and Elufioye, Oluwafunmi Adijat and Olorunsogo, Temidayo and Asuzu, Onyeka Franca and Nduubuisi, Ndubuisi Leonard and Daraojimba, Andrew Ifesinachi},
  journal={World Journal of Advanced Research and Reviews},
  volume={21},
  number={02},
  pages={1750--1760},
  year={2024}
}

@article{wang2025linkalign,
  title={LinkAlign: Scalable Schema Linking for Real-World Large-Scale Multi-Database Text-to-SQL},
  author={Wang, Yihan and Liu, Peiyu},
  journal={arXiv preprint arXiv:2503.18596},
  year={2025}
}

@inproceedings{cen2025sqlfixagent,
  title={SQLFixAgent: Towards Semantic-Accurate Text-to-SQL Parsing via Consistency-Enhanced Multi-Agent Collaboration},
  author={Cen, Jipeng and Liu, Jiaxin and Li, Zhixu and Wang, Jingjing},
  booktitle={Proceedings of the AAAI Conference on Artificial Intelligence},
  volume={39},
  number={1},
  pages={49--57},
  year={2025}
}

@article{zhou2023learned,
  title={A learned query rewrite system},
  author={Zhou, Xuanhe and Li, Guoliang and Wu, Jianming and Liu, Jiesi and Sun, Zhaoyan and Zhang, Xinning},
  journal={Proceedings of the VLDB Endowment},
  volume={16},
  number={12},
  pages={4110--4113},
  year={2023},
  publisher={VLDB Endowment}
}

@misc{openai2024code,
  author       = {{OpenAI}},
  title        = {Code Interpreter},
  year         = {2024},
  howpublished = {\url{https://platform.openai.com/docs/assistants/tools/code-interpreter}},
  note         = {Accessed: 2025-08-04}
}

@misc{openai2024dataanalysis,
  author       = {{OpenAI}},
  title        = {Data Analysis with Chatgpt},
  year         = {2024},
  howpublished = {\url{https: //help.openai.com/en/articles/ 8437071-data-analysis-with-chatgpt}},
  note         = {Accessed: 2025-08-04}
}

@inproceedings{vacareanu2024words,
title={From Words to Numbers: Your Large Language Model Is Secretly A Capable Regressor When Given In-Context Examples},
author={Robert Vacareanu and Vlad Andrei Negru and Vasile Suciu and Mihai Surdeanu},
booktitle={First Conference on Language Modeling},
year={2024},
url={https://openreview.net/forum?id=LzpaUxcNFK}
}

@inproceedings{cheng2023gpt,
  title={Is GPT-4 a Good Data Analyst?},
  author={Cheng, Liying and Li, Xingxuan and Bing, Lidong},
  booktitle={Findings of the Association for Computational Linguistics: EMNLP 2023},
  pages={9496--9514},
  year={2023}
}

@inproceedings{ma-etal-2023-insightpilot,
    title = "{I}nsight{P}ilot: An {LLM}-Empowered Automated Data Exploration System",
    author = "Ma, Pingchuan  and
      Ding, Rui  and
      Wang, Shuai  and
      Han, Shi  and
      Zhang, Dongmei",
    editor = "Feng, Yansong  and
      Lefever, Els",
    booktitle = "Proceedings of the 2023 Conference on Empirical Methods in Natural Language Processing: System Demonstrations",
    month = dec,
    year = "2023",
    address = "Singapore",
    publisher = "Association for Computational Linguistics",
    url = "https://aclanthology.org/2023.emnlp-demo.31/",
    doi = "10.18653/v1/2023.emnlp-demo.31",
    pages = "346--352"
}

@inproceedings{wang2023plan,
    title = "Plan-and-Solve Prompting: Improving Zero-Shot Chain-of-Thought Reasoning by Large Language Models",
    author = "Wang, Lei  and
      Xu, Wanyu  and
      Lan, Yihuai  and
      Hu, Zhiqiang  and
      Lan, Yunshi  and
      Lee, Roy Ka-Wei  and
      Lim, Ee-Peng",
    editor = "Rogers, Anna  and
      Boyd-Graber, Jordan  and
      Okazaki, Naoaki",
    booktitle = "Proceedings of the 61st Annual Meeting of the Association for Computational Linguistics (Volume 1: Long Papers)",
    month = jul,
    year = "2023",
    address = "Toronto, Canada",
    publisher = "Association for Computational Linguistics",
    url = "https://aclanthology.org/2023.acl-long.147/",
    doi = "10.18653/v1/2023.acl-long.147",
    pages = "2609--2634"
}

@inproceedings{wang2024executable,
  title={Executable code actions elicit better llm agents},
  author={Wang, Xingyao and Chen, Yangyi and Yuan, Lifan and Zhang, Yizhe and Li, Yunzhu and Peng, Hao and Ji, Heng},
  booktitle={Forty-first International Conference on Machine Learning},
  year={2024}
}

@inproceedings{hong2024data,
    title = "Data Interpreter: An {LLM} Agent for Data Science",
    author = "Hong, Sirui  and
      Lin, Yizhang  and
      Liu, Bang  and
      Liu, Bangbang  and
      Wu, Binhao  and
      Zhang, Ceyao  and
      Li, Danyang  and
      Chen, Jiaqi  and
      Zhang, Jiayi  and
      Wang, Jinlin  and
      Zhang, Li  and
      Zhang, Lingyao  and
      Yang, Min  and
      Zhuge, Mingchen  and
      Guo, Taicheng  and
      Zhou, Tuo  and
      Tao, Wei  and
      Tang, Robert  and
      Lu, Xiangtao  and
      Zheng, Xiawu  and
      Liang, Xinbing  and
      Fei, Yaying  and
      Cheng, Yuheng  and
      Ni, Yongxin  and
      Gou, Zhibin  and
      Xu, Zongze  and
      Luo, Yuyu  and
      Wu, Chenglin",
    editor = "Che, Wanxiang  and
      Nabende, Joyce  and
      Shutova, Ekaterina  and
      Pilehvar, Mohammad Taher",
    booktitle = "Findings of the Association for Computational Linguistics: ACL 2025",
    month = jul,
    year = "2025",
    address = "Vienna, Austria",
    publisher = "Association for Computational Linguistics",
    url = "https://aclanthology.org/2025.findings-acl.1016/",
    doi = "10.18653/v1/2025.findings-acl.1016",
    pages = "19796--19821",
    ISBN = "979-8-89176-256-5"
}

@inproceedings{ding2019quickinsights,
  title={Quickinsights: Quick and automatic discovery of insights from multi-dimensional data},
  author={Ding, Rui and Han, Shi and Xu, Yong and Zhang, Haidong and Zhang, Dongmei},
  booktitle={Proceedings of the 2019 international conference on management of data},
  pages={317--332},
  year={2019}
}

@inproceedings{law2020characterizing,
  title={Characterizing automated data insights},
  author={Law, Po-Ming and Endert, Alex and Stasko, John},
  booktitle={2020 IEEE Visualization Conference (VIS)},
  pages={171--175},
  year={2020},
  organization={IEEE}
}

@inproceedings{hong2024knowledge,
  title={Knowledge-to-SQL: Enhancing SQL Generation with Data Expert LLM},
  author={Hong, Zijin and Yuan, Zheng and Chen, Hao and Zhang, Qinggang and Huang, Feiran and Huang, Xiao},
  booktitle={Findings of the Association for Computational Linguistics ACL 2024},
  pages={10997--11008},
  year={2024}
}

\onecolumn
\newpage
\twocolumn
\appendix

\begin{table*}[t!]
\centering
\begin{tabular}{>{\justifying\arraybackslash}m{3cm}>{\justifying\arraybackslash}m{3cm}>{\justifying\arraybackslash}m{4cm}>{\justifying\arraybackslash}m{4cm}}
\toprule
Error Type & Analysis Task & AgentPoirot Output & Correct Output \\
\midrule
\noindent Underutilization of Domain Knowledge & \noindent Analyze regional sales fluctuations & \noindent In early February, sales in northern China dropped sharply. Due to the lack of additional information in the data, \textcolor{myred}{\textbf{this was attributed to random variance.}} & \noindent \textcolor{mygreen}{\textbf{The drop actually aligns with the Chinese Spring Festival,}} a major national holiday that significantly disrupts commercial activity.\\
\midrule
\noindent Shallow Analytical Depth & \noindent Diagnose a sudden drop in gross margin & \noindent \textcolor{myred}{\textbf{Raise shallow question:}} ``Which category has the highest cost increase?''
 & \noindent \textcolor{mygreen}{\textbf{Insightful Question:}} ``Is the margin drop concentrated in specific provinces or time periods?''
 \\
\midrule
\noindent Error-Prone Code Generation & \noindent Evaluate the impact of a summer marketing campaign on sales & \noindent Join a sales table (with a date column in string format) with a marketing table where the date column is stored as a datetime object. -> \textcolor{myred}{\textbf{Empty result set.}}
 & \noindent \textcolor{mygreen}{\textbf{Cast the types properly before joining.}} \\
\bottomrule
\end{tabular}
\caption{Failure cases of the state-of-the-art \textit{AgentPoirot} model in real-world insight discovery tasks. We categorize these errors into three major types: \textbf{domain knowledge underutilization}, \textbf{shallow analytical depth}, and \textbf{error-prone code generation}. \textcolor{myred}{\textbf{Red}} text highlights the model's mistakes, while \textcolor{mygreen}{\textbf{green}} indicates correct output.}
\label{tab:error_analysis}
\end{table*}

\section{Error Analysis: Failures of Existing Data Agents}\label{app:error_analysis}
Despite recent advancements in LLM-based data agents, we observe consistent and critical limitations when these systems are applied to real-world insight discovery tasks. As shown in Table~\ref{tab:error_analysis}, we categorize these limitations into three major types: domain knowledge underutilization, shallow analytical depth, and error-prone code generation. Through empirical observations on the state-of-the-art AgentPoirot model~\cite{sahu2025insightbench}, we identify and analyze representative failure cases.

\subsection{Underutilization of Domain Knowledge}
Most existing agents~\cite{perez2025llm,sahu2025insightbench} assume that relevant knowledge is either encoded within the LLM's parameters or explicitly provided by the user. In practice, however, many analytical tasks require external domain context (such as definitions of business KPIs, seasonal market patterns, or industry-specific thresholds), which are rarely captured by LLMs.
For example, when analyzing regional sales fluctuations, AgentPoirot observes a sharp drop in sales in northern China in early February and concludes that it is random variance. However, this drop actually aligns with the Chinese Spring Festival, a major national holiday that significantly disrupts commercial activity. Without this domain-specific knowledge, the agent fails to recognize the true cause, leading to a misleading conclusion.
In another case, when analyzing marketing campaign effectiveness, AgentPoirot flags a low conversion rate as problematic without realizing that the campaign is intentionally targeted for long-term brand building rather than short-term sales.
These examples show that without targeted domain augmentation or external retrieval mechanisms, agents struggle to distinguish signal from noise in context-rich data, often producing superficial or misleading insights.

\subsection{Shallow Analytical Depth}
Current systems~\cite{sahu2025insightbench,wu2024automated,langchain2024pandas} often rely on a single-pass LLM to generate analytical questions based on dataset description or metadata. While this setup can yield surface-level questions (such as identifying a top-selling product or calculating simple correlations), it struggles to generate deeper, multi-layered questions that require structured reasoning and contextual understanding. For instance, in a diagnostic task involving a sudden drop in gross margin, AgentPoirot asks, ``Which category has the highest cost increase?'', but fails to propose more insightful questions such as ``Is the margin drop concentrated in specific provinces or time periods?'' or ``Do seasonal campaigns correlate with shifts in cost structure?''. This lack of depth originates from the absence of iterative, divergent-convergent thinking processes, and limits the agent’s ability to guide users toward non-obvious, high-value insights in complex domains like finance, healthcare, or operations.

\subsection{Error-Prone Code Generation}
LLMs are known to hallucinate or generate incorrect code~\cite{huynh2025large, hong2024next}, especially in non-trivial scenarios involving intermediate state reuse, complex data joins, or edge case handling. We observe some recurring failure modes:
(1) Semantically correct but logically irrelevant code. For example, given a question like ``Which customer segments showed declining profitability over the last two quarters'',  AgentPoirot instead computes overall revenue changes across all customers, missing the segment-level breakdown entirely.
(2) Lack of robustness to schema inconsistencies. A common issue occurs when joining tables with mismatched column formats. In one case, AgentPoirot attempts to join a sales table (with a date column in string format like `2023-07-01') with a marketing table where the date column is stored as a datetime object. The failure to cast the types properly causes the join to return an empty result set, leading to misleading ``no impact'' conclusions.
Even when users provide detailed prompts and table descriptions, the generated code often lacks defensive programming practices such as type checking or exception handling. These errors significantly reduce trust in autonomous agents for insight generation.

These failure patterns highlight a fundamental gap between existing data agents’ capabilities and the practical demands of robust insight discovery. They motivate our design of a multi-agent architecture that explicitly addresses these shortcomings by incorporating domain-aware knowledge retrieval, divergent-convergent multi-role debating question raising, and multi-path reasoning for code generation.

\section{More Experimental Results}
\label{app:experimental_results}
\subsection{Ablation Study}

\subsubsection{The effect of RAKG module}

\begin{table*}[ht]
\centering
\begin{tabular}{lccc}
\toprule
\textbf{Model Variant} & \textbf{G-Eval (Insight)} & \textbf{G-Eval (Summary)} & \textbf{Retrieval Usage Rate (\%)} \\
\midrule
No Retrieval & 0.3316 & 0.3982 & 0\% \\
Internal Knowledge & 0.3366 & 0.3994 & 0\% \\
On-Demand Retrieval & 0.3530 & 0.4059 & 24\% \\
Full Retrieval & 0.3560 & 0.4082 & 100\% \\
\bottomrule
\end{tabular}
\caption{
Performance comparison of four different retrieval strategies in~\modelname. Our on-demand retrieval mechanism strikes an effective balance between efficiency and performance, achieving near-optimal performance while significantly reducing the search resource consumption compared to full retrieval.
}
\label{tab:knowledge_search_comparison}
\end{table*}

To demonstrate the effectiveness of our RAKG module design, we conduct a comprehensive comparison of different retrieval strategies within our framework.
We compare four different variants: No Retrieval (completely without any retrieval), Internal Knowledge (using \textit{vanilla knowledge generator} leveraging only the internal knowledge of LLMs), On-Demand Retrieval (used in our RAKG module), and Full Retrieval (retrieval is utilized for all analysis tasks).
As shown in Table~\ref{tab:knowledge_search_comparison}, there are several interesting findings worth highlighting.

First, employing a \textit{vanilla knowledge generator} yields a modest improvement over the no retrieval baseline. It indicates that the \textit{vanilla knowledge generator} can further stimulate the latent knowledge encoded within LLMs, which is also consistent with prior works~\cite{zhang2025data,hong2024knowledge}. It validates our design choice of incorporating the \textit{vanilla knowledge generator} in RAKG module.

Second, integrating external retrieval significantly enhances model performance, demonstrating the critical role of external knowledge in real-world data analysis scenarios.

Third, among the four variants, the full retrieval approach achieves the highest scores, yet at the cost of invoking search queries for all instances, which substantially increases resource demands.
In contrast, the on-demand retrieval mechanism strikes an effective balance between efficiency and performance. Despite utilizing only 24\% of the search resources required by the full retrieval method, it achieves comparable performance. It suggests that on-demand retrieval can reduce resource consumption without compromising model effectiveness, highlighting the practical advantages of our dynamic retrieval design.

\subsubsection{The effect of Question Raising module}


\modelname~employs a divergent-convergent multi-role debating process in Question Raising module, aiming to simulate diverse analytical perspectives and deepen the analytical depth. To experimentally validate this design, we compare the questions raised by our framework with those from the best baseline, AgentPoirot, evaluating both diversity and coverage of the raised questions.

We quantify diversity as the average pairwise dissimilarity between question embeddings, calculated by: 
\begin{equation}
\text{Diversity} = 1 - \frac{2}{n(n-1)} \sum_{i<j} \text{cosine\_sim}(v_i, v_j),
\end{equation}
where $v_i$ and $v_j$ are the embedding vectors of questions $i$ and $j$. Coverage is defined as the average radius of the questions around their centroid embedding, computed by: 
\begin{equation}
\text{Coverage} = \frac{1}{n} \sum_i \| v_i - \text{centroid} \|,
\end{equation}
where the centroid is the mean vector of all question embeddings.

\begin{figure}[t!]
    \centering
    \includegraphics[width=0.95\linewidth]{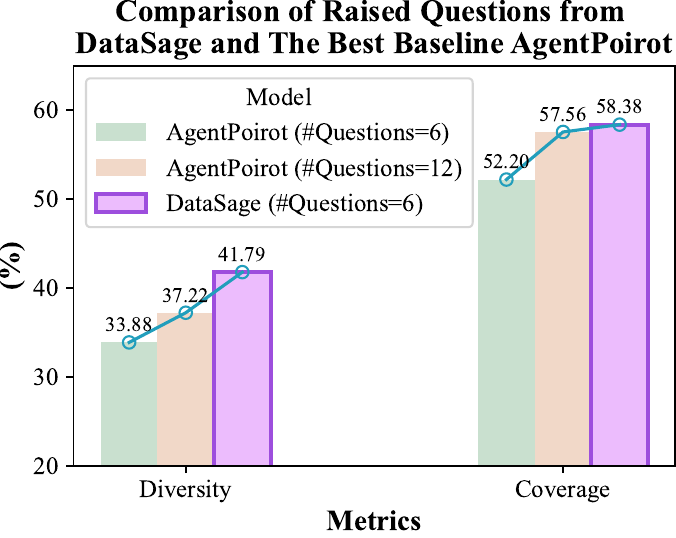}
    \caption{Comparison of diversity and coverage of the questions raised by~\modelname~and the best baseline AgentPoirot. Even when generating fewer questions (6 vs. 12), \modelname~produces questions with significantly higher diversity and coverage, enabling the generation of deeper and more diverse insights.
    }
    \label{fig:diversity_coverage}
\end{figure}

As shown in Figure~\ref{fig:diversity_coverage}, \modelname~achieves significantly higher diversity and coverage scores compared to AgentPoirot, even when generating fewer
questions. Although increasing the number of raised questions generally improves these metrics, AgentPoirot's questions tend to be more similar or even repetitive, limiting its diversity and coverage of the analytical space.
This often leads to the generation of shallow and generic insights, whereas~\modelname~effectively promotes the creation of diverse and comprehensive questions, enabling the generation of deeper and more diverse insights.

\subsubsection{The effect of Multi-path Reasoning}

\begin{table}[t!]
\centering
\begin{tabular}{lcc}
\toprule
\textbf{Model} & \textbf{Success Rate} & \textbf{\#Refinements} \\
\midrule
AgentPoirot  & 95.17\% & - \\
\midrule
\modelname & \textbf{99.50\%} & \textbf{1.36} \\
\quad w/o MR & 98.17\% & 1.63 \\
\bottomrule
\end{tabular}
\caption{The effect of Multi-path Reasoning (MR) on code execution success rate and average number of code refinements. Multi-path Reasoning not only improves code correctness but also reduces refinement overhead.}
\label{tab:MR}
\end{table}
To further examine the benefits of our proposed Multi-path Reasoning (MR), we evaluate its impact on both code correctness and refinement efficiency. Specifically, we use a simple yet effective proxy for correctness: the execution success rate, i.e., the proportion of generated code snippets that run without errors. In addition, we measure the average number of code refinements required in the Insight Generation Module. We compare three variants:~\modelname,~\modelname~without MR, and the best baseline AgentPoirot.

As shown in Table~\ref{tab:MR},~\modelname~significantly outperforms the best baseline, achieving a 4.33 percentage point gain in execution success rate (99.50\% vs. 95.17\%). Moreover, comparing~\modelname~to its variant without MR highlights the effectiveness of Multi-path Reasoning. The success rate improves from 98.17\% to 99.50\%, while the average number of refinements decreases from 1.63 to 1.36. This indicates that Multi-path Reasoning can not only increase code correctness but also reduce the reliance on post-hoc refinement. It enhances robustness by mitigating single-path reasoning errors, while also improving efficiency by reducing unnecessary refinement iterations.

\begin{figure}[t!]
    \centering
    \includegraphics[width=0.95\linewidth]{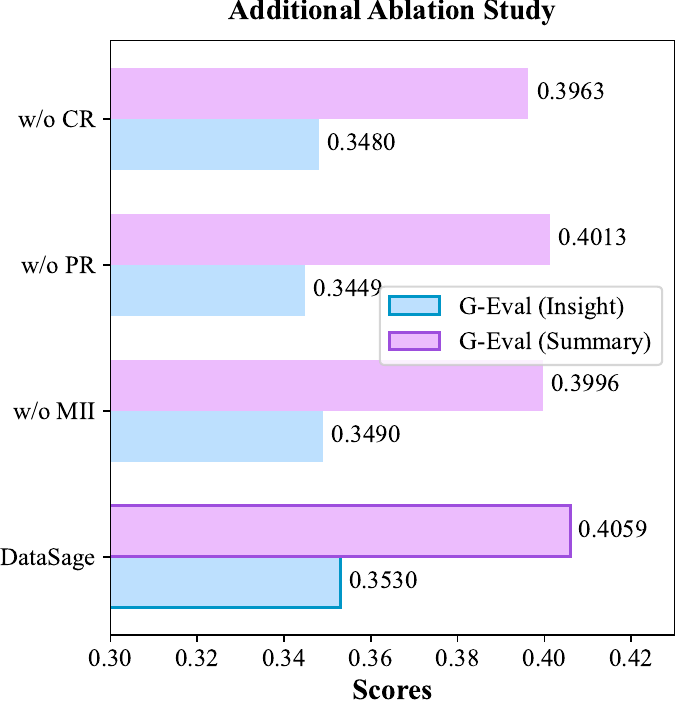}
    \caption{Additional ablation study results. Removing Multimodal Insight Interpretation (MII), Plot Reviewer (PR), or Code Refinement (CR) consistently degrades performance, confirming the necessity of these components.
}
    \label{fig:ablation_app}
\end{figure}

\subsubsection{Additional Ablation Study}
Beyond the core components of~\modelname, we further investigate the impact of several finer-grained design choices. Specifically, we examine three variants: (i) removing Multimodal Insight Interpretation (using only text-level interpretation), (ii) removing the Plot Reviewer, and (iii) removing Code Refinement. As shown in Figure~\ref{fig:ablation_app}, the absence of any of these components consistently degrades performance. This validates the necessity of our full design. Notably, Multimodal Insight Interpretation and Plot Reviewer are complementary. The Plot Reviewer enhances the quality of generated plots, thereby providing more reliable inputs for multimodal interpretation. Meanwhile, Code Refinement improves the correctness of generated code, which directly contributes to more accurate insight discovery. These results highlight the synergistic contributions of different components and demonstrate the robustness of our framework.

\subsection{Hyperparameter Study}

\begin{figure}[t!]
    \centering
    \includegraphics[width=\linewidth]{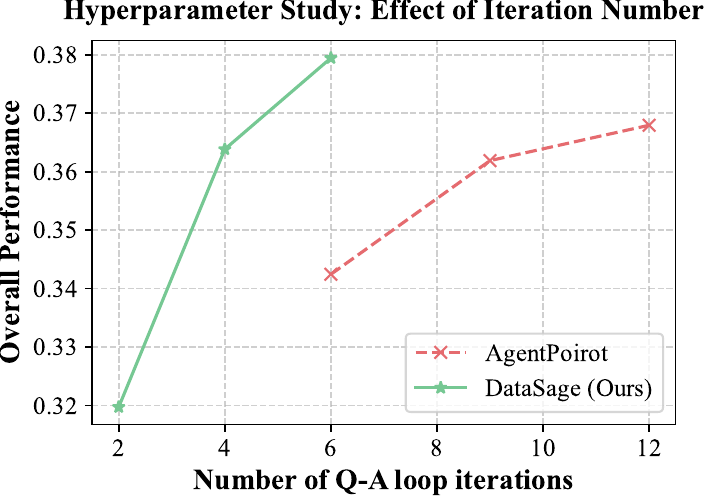}
    \caption{Hyperparameter study on $N_{\text{iter}}$ (the number of Q-A loop iterations). Both models generally benefit from more iterations. \modelname~achieves competitive or superior performance with significantly fewer iterations compared to the baseline.}
    \label{fig:plot_hyper}
\end{figure}

We further investigate the impact of the most critical hyperparameter, $N_{\text{iter}}$, which controls the number of Q-A loop iterations of the framework.
As shown in Figure~\ref{fig:plot_hyper}, we report the overall performance of our framework~\modelname~and the baseline AgentPoirot under different values of $N_{\text{iter}}$. For both methods, performance generally improves as the number of iterations increases, which suggests that more interations enable the framework to better discovery insights. However, \modelname~achieves comparable or even superior results with significantly fewer iterations. In particular, \modelname~already surpasses the baseline with 9 iterations when using only 4 iterations. This demonstrates that our design is not only more effective but also more efficient, requiring fewer iterations to reach strong performance.
Overall, the results confirm that increasing the iteration budget is beneficial, but the gains quickly saturate for baseline model. In contrast, \modelname~leverages its structured design to extract higher-quality insights with substantially fewer iterations, highlighting the efficiency advantage of our framework.

\subsection{Additional Experimental Analysis}
\subsubsection{Multi-path Reasoning Statistics}

\begin{figure}[ht]
    \centering
    \includegraphics[width=0.95\linewidth]{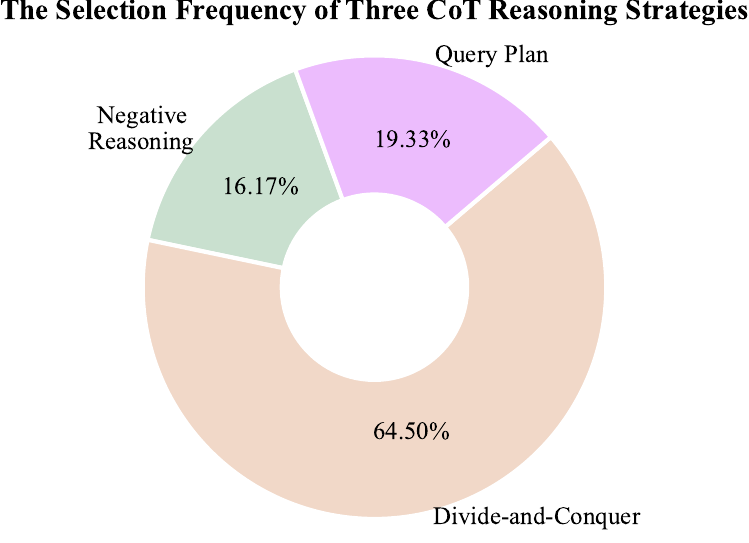}
    \caption{
    Distribution of the final selected code outputs generated by three CoT reasoning strategies. DaC accounts for the majority of selections (64.50\%), highlighting its effectiveness in data analysis tasks, while QP and NR contribute complementary reasoning capabilities that enhance overall code correctness.
    }
    \label{fig:multipath_statistics}
\end{figure}

We conduct a statistical analysis on the selection frequency of three complementary CoT reasoning strategies (i.e., Divide-and-Conquer, Query Plan, and Negative Reasoning). As shown in Figure~\ref{fig:multipath_statistics}, the DaC (Divide-and-Conquer) strategy is the most frequently chosen, accounting for 64.50\% of the final selected code outputs, followed by QP (Query Plan) at 19.33\%, and NR (Negative Reasoning) at 16.17\%. This distribution suggests that in data analysis scenarios, the DaC approach tends to be more effective and reliable in generating correct code. Although the selection rates for QP and NR are relatively lower, these strategies play a complementary role, enabling the model to leverage the strengths of different reasoning paths and ultimately select the most accurate solution. This complementary interplay increases the overall likelihood of producing correct code. The statistical results also motivate us to prioritize the DaC strategy for code generation in resource-constrained environments.

\section{Case Study}
\label{app:case_study}

\begin{table*}[t!]
\centering
\small
\begin{tabular}{p{4.5cm} p{4.2cm} p{5.5cm}}
\toprule
\textbf{Insights Predicted by AgentPoirot (The Best Baseline Model)} & \textbf{Advantages of~\modelname} & \textbf{Insights Predicted by~\modelname~(Ours)} \\
\midrule
\multicolumn{3}{c}{\textbf{Task: \textit{Incidents Management}}} \\
\midrule
The number of incident tickets created by ``David Loo'' nearly doubled over the span of one year, rising from 16 incidents in early 2023 to 32 incidents in early 2024, highlighting a significant increase in reported issues from this single caller which may indicate growing system problems or increased user activity. &
Adds slope and significance testing for trend validation, and incorporates temporal distribution analysis (hourly, weekly), enhancing both rigor and coverage. \textbf{\textit{[Statistical Quantification]}} & 
David Loo’s incident ticket submissions increased by approximately 100\% over the year, with a significant positive monthly trend \textbf{(slope = 0.057, p $\mathbf{\approx}$ 1.25e-5)}, and his ticket volume remains consistently higher than other callers \textbf{across all hours and weekdays}, highlighting a persistent and growing source of incidents that may require focused attention. \\
\midrule
David Loo's incident ticket creation nearly doubled over the span of a year, highlighting a growing demand or issue source from this caller that may require targeted investigation. &
Links growth to duplicate submissions, quantifies rate change, and identifies temporal peaks, enabling targeted root cause investigation. \textbf{\textit{[Root Cause Analysis]}} &
Among all callers, ``David Loo'' uniquely exhibits a measurable upward trend in submitting duplicate or near-duplicate incident tickets, with the number of such tickets increasing by approximately 0.0032 per day (about 1 extra duplicate every 10 months), particularly between September and November 2023, \textbf{suggesting that his reported incident volume is partially driven by repeated submissions of similar problems within short timeframes}. \\
\midrule
The data reveals a growing trend in high-priority incidents reported by David Loo, with ``2 - High'' priority incidents more than doubling over the year, highlighting a potential escalation in the severity or frequency of issues faced by this caller. &
Extends priority-level analysis with temporal clustering metrics (97\%) and category detail (network), providing precise basis for emergency planning. \textbf{\textit{[Operational Planning]}} &
David Loo’s incident reports not only increase sharply by nearly two incidents per month but also cluster tightly in time, with 97\% occurring within a week of each other, mostly involving high-priority network issues, \textbf{suggesting urgent, recurring problems that may require targeted operational attention}. \\
\midrule
Despite the ITIL User having increasing incident reports, their priority distribution remains aligned with other callers, with approximately 75\% of their incidents classified as ``2 - High,'' suggesting that the increase is not driven by a shift toward more severe or less severe incident priorities. &
Shifts focus from static priority comparison to operational efficiency and workload impact, offering direct guidance for staffing decisions. \textbf{\textit{[Resource Allocation]}} &
Despite a similar average resolution time (175 vs. 177 hours), callers with rising incident volumes cause a substantial increase in workload, with incident counts assigned to each support staff member nearly doubling (e.g., from 27--40 to 59--88 incidents), highlighting that \textbf{growing incident frequency demands more extensive support resource allocation without degrading resolution speed}. \\
\midrule
While ITIL User reports fewer incidents overall, about 69\% (69 out of 75) of their incidents are high or critical priority, indicating that their reported incidents are relatively more severe compared to the distribution of priorities from other callers. &
Integrates severity with lifecycle analysis, detecting reopen loops and processing delays, extending severity analysis into process stability evaluation. \textbf{\textit{[Process Analysis]}} &
Callers with rising incident volumes, notably ``David Loo'' and ``Don Goodliffe,'' have average incident open durations ranging from approximately 140 to over 310 hours monthly, with 12 to 13 months showing overlapping ``Resolved'' and ``Closed'' states that suggest frequent incident reopenings, indicating \textbf{persistent delays and instability in incident resolution that likely impact service efficiency and warrant targeted process improvements}. \\
\bottomrule
\end{tabular}
\caption{
A case study comparing the insight outputs from our framework~\modelname~and the best baseline model AgentPoirot. While AgentPoirot detects broad patterns, it remains at a descriptive level and often lacks statistical rigor, in-depth analysis, and operational context. The advantages of~\modelname~over the baseline model are highlighted in \textbf{bold}, including statistically validated, multi-faceted, and more actionable insights, which offer more robust decision support.
}
\label{tab:case_study}
\end{table*}

\subsection{\modelname~generates better insights.}
To better demonstrate the capability of~\modelname~in generating actionable insights, we conduct a case study on an incidents management task in InsightBench. As shown in Table~\ref{tab:case_study}, we compare the outputs of our framework with those from the best baseline model AgentPoirot. The advantages of our framework can be summarized in the following three points.

\noindent\textbf{Statistically Validated Insights.}\quad
The baseline model identifies the general growth trend in David Loo’s incident submissions, reporting that the count nearly doubled over the year and highlighting an increase in high-priority incidents. However, these descriptions remain at a descriptive level and lack supporting statistical evidence or precise quantification. In contrast, \modelname~provides quantitative trend metrics, such as slope estimates (slope = 0.057, p $\mathbf{\approx}$ 1.25e-5), explicit percentage increases ($\mathbf{\approx}$100\%), and temporal clustering statistics (97\% within one week). These quantitative markers not only confirm the trend but also allow for statistical significance assessment.

\noindent\textbf{Multi-Dimensional In-Depth Insights.}\quad
While the baseline model’s focus is largely confined to incident counts and priority levels, \modelname~extends the analysis to multiple operational dimensions:
\begin{itemize}
    \setlength{\itemsep}{0em}
\item Resolution efficiency: Comparing average resolution times (175 vs. 177 hours) while identifying resource strain due to increased workload per staff member.
\item Duplicate submissions: Detecting a measurable upward trend in duplicate or near-duplicate incident tickets (0.0032/day), particularly concentrated in specific months.
\item Process inefficiencies: Highlighting frequent reopenings of tickets based on overlapping ``Resolved'' and ``Closed'' states.
\end{itemize}
This is attributed to our carefully designed Question Raising Module, which can simulate diverse analytical perspectives and deepen analytical depth.

\noindent\textbf{More Actionable Insights.}\quad
The baseline model remains limited to high-level descriptions and does not capture underlying process issues or secondary contributing factors. Our framework effectively bridges this gap by producing operationally actionable insights that enable targeted, data-driven decision-making:
\begin{itemize}
    \setlength{\itemsep}{0em}
\item Pinpointing peak months and weeks for incident clustering, enabling targeted root cause investigation.
\item Quantifying workload per support staff member to guide resource allocation.
\item Identifying increasing duplicate or near-duplicate ticket submissions, along with frequent incident reopenings, providing a clear direction for process improvements.
\end{itemize}

In summary, the case study demonstrates that while the best baseline model AgentPoirot can detect broad patterns, it often stops at a descriptive level, lacking statistical rigor, in-depth analysis and operational context. Our framework enriches them with statistically validated, multi-faceted, and actionable insights, thereby offering more robust decision support for organizations.

\subsection{\modelname~generates better plots.}
\label{app:case_study_plot}

\begin{table*}[t!]
\centering
\small
\begin{tabular}{m{5.5cm} m{3.6cm} m{5.5cm}}
\toprule
\textbf{Plots Generated by AgentPoirot (The Best Baseline Model)} & \textbf{Advantages of~\modelname} & \textbf{Plots Generated by~\modelname~(Ours)} \\
\midrule
\multicolumn{3}{c}{\textbf{Task: \textit{Incidents Management}}} \\
\midrule
\includegraphics[width=5.5cm]{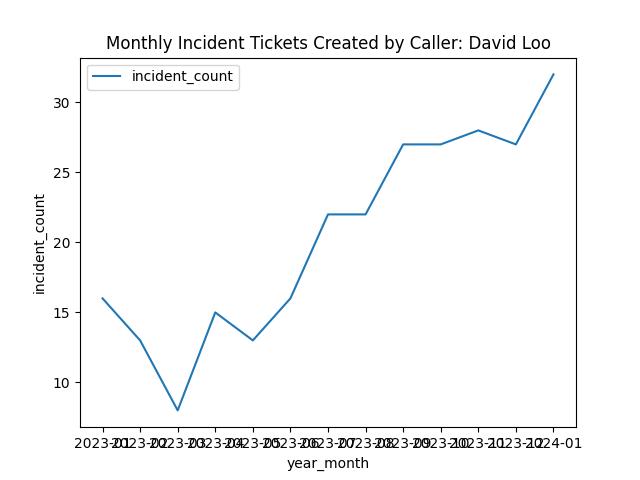} & Introduces group-level comparisons to contextualize individual trends, identifying unique vs. systemic patterns to enhance decision relevance. \textbf{\textit{[More comparative and informative]}} & \includegraphics[width=5.5cm]{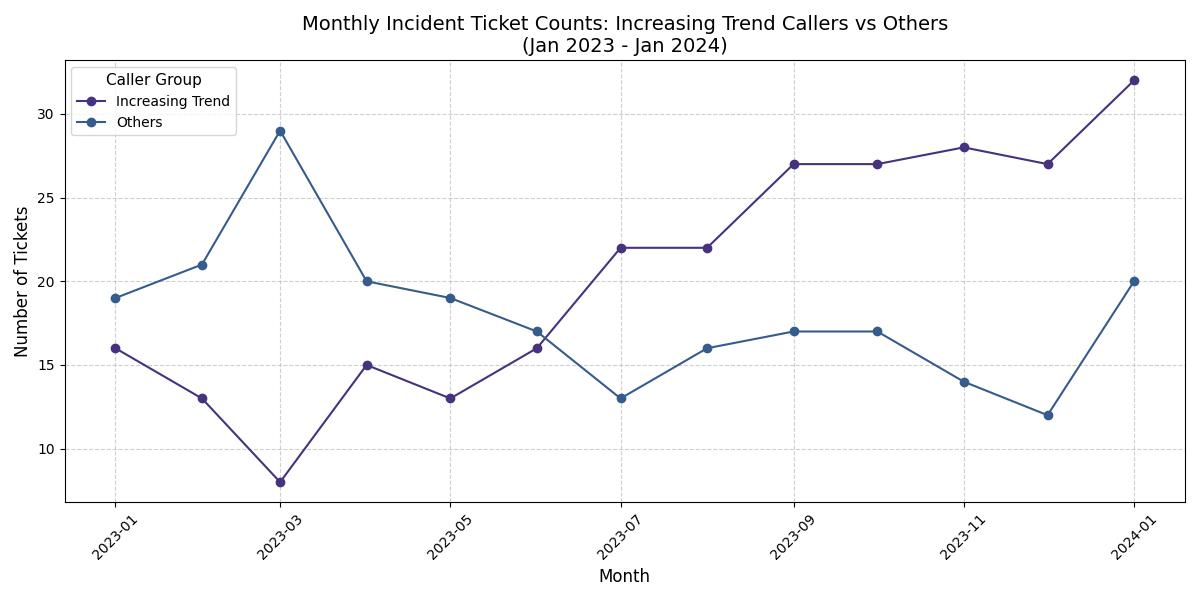} \\
\midrule
\includegraphics[width=5.5cm]{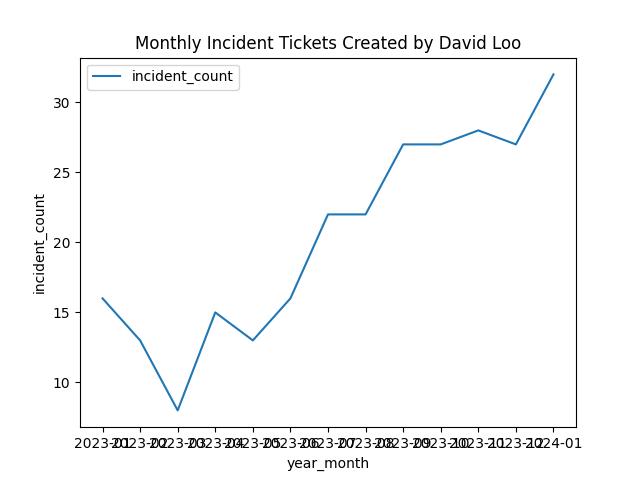} & Generates diversified, non-redundant plots and insights, linking growth to duplicate submissions for better targeted investigation. \textbf{\textit{[More diverse and non-redundant]}} & \includegraphics[width=5.5cm]{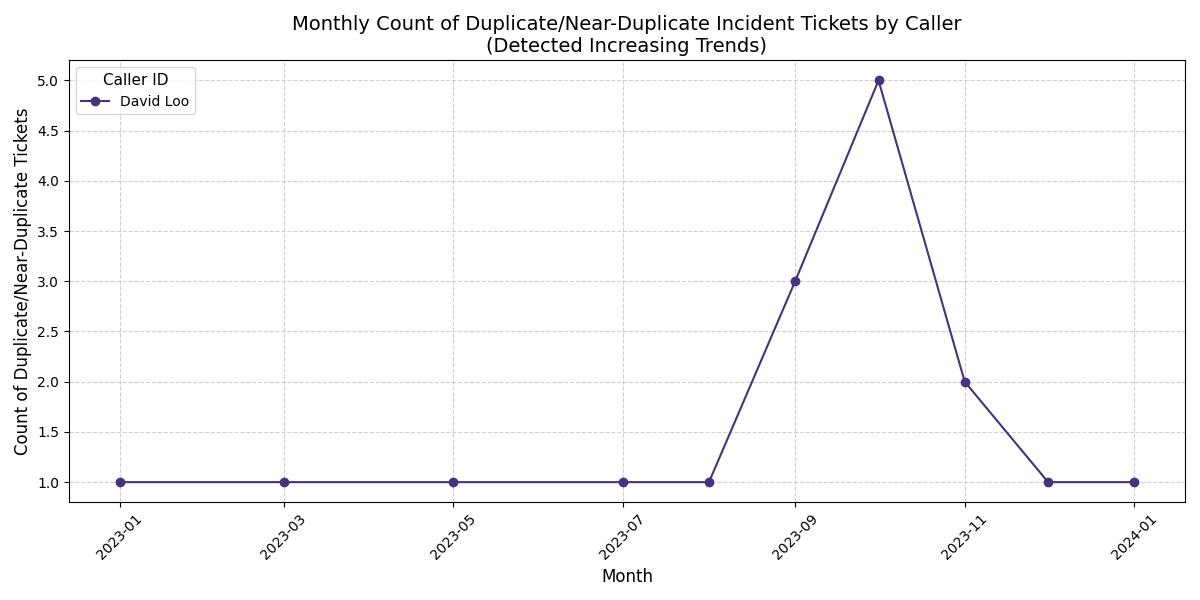}\\
\midrule
\includegraphics[width=5.5cm]{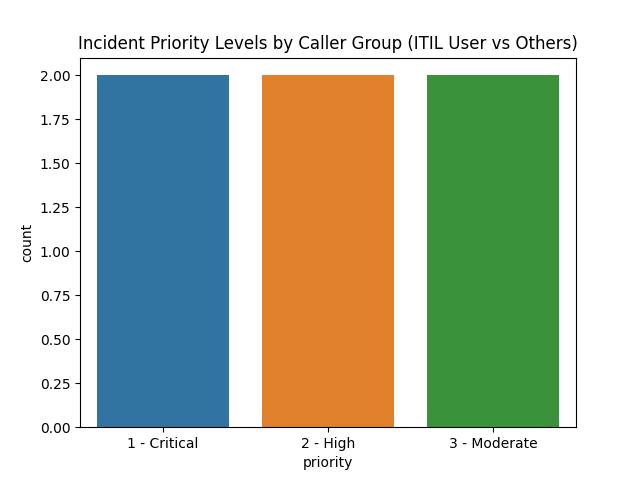} & Generates correct plots, fixing baseline’s coding error (all bars have the same height), and shifts focus to operational efficiency and workload impact for staffing decisions. \textbf{\textit{[Error-free code and plot generation]}} & \includegraphics[width=5.5cm]{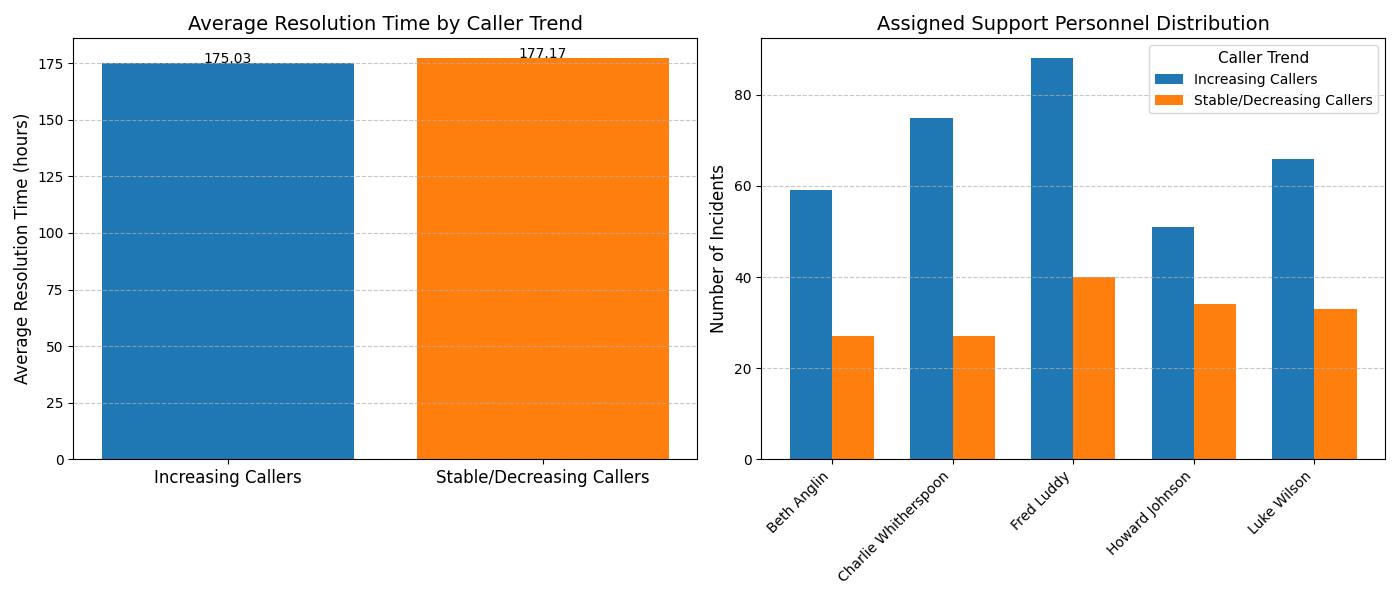} \\
\midrule
\includegraphics[width=5.5cm]{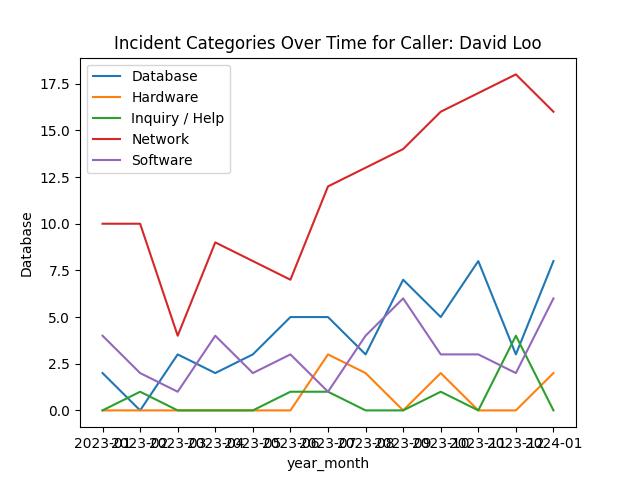} & Replaces baseline’s cluttered, mislabeled (y-axis titled ``Database'' instead of ``Incident Count'') line chart with a clear stacked-area design, removing overlap and simplifying time labels for more interpretable plots. \textbf{\textit{[More readable and interpretable]}} & \includegraphics[width=5.5cm]{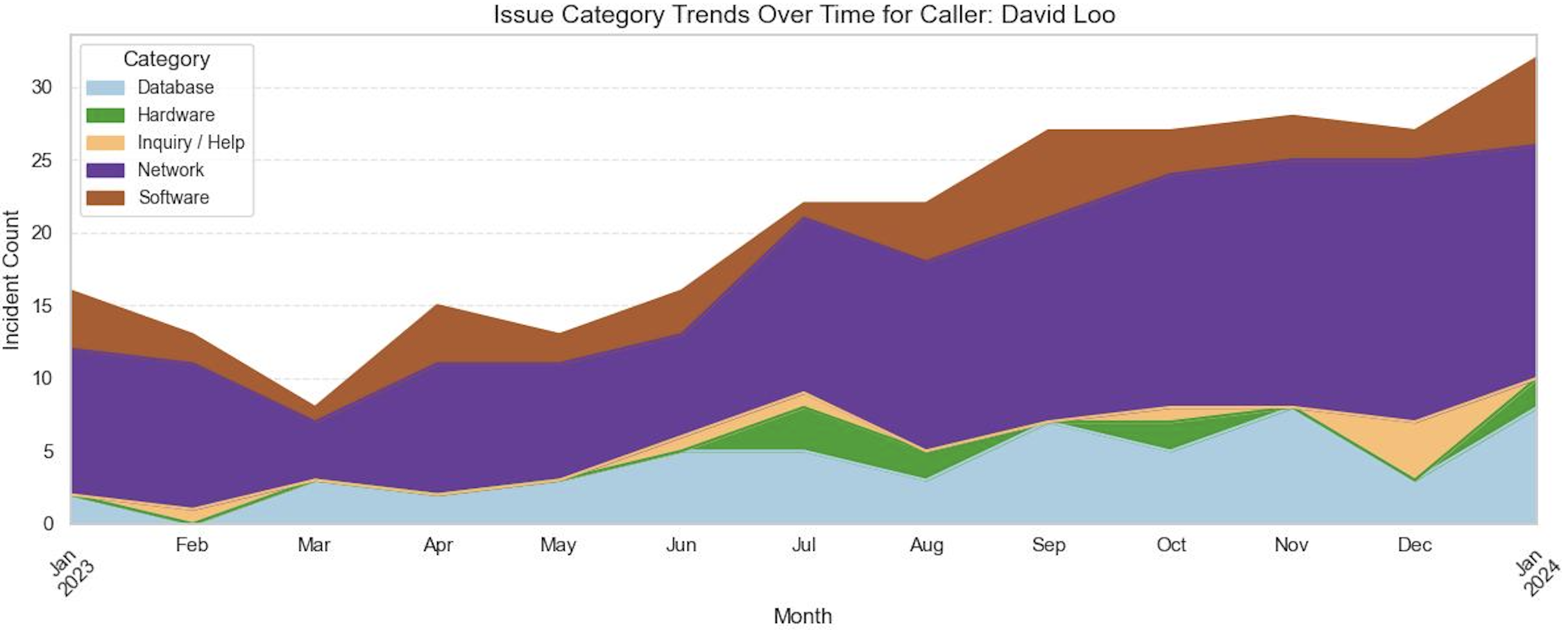} \\
\bottomrule
\end{tabular}
\caption{
A case study comparing the plots generated from our framework~\modelname~and the best baseline model AgentPoirot. While AgentPoirot can produce basic visualizations, its plots tend to be narrowly focused, redundant, error-prone, and suffer from poor readability. In contrast, \modelname~demonstrates clear advantages including richer information, more diverse and non-redundant plots, error-free plot generation, and enhanced readability and interpretability.
}
\label{tab:case_study_plot}
\end{table*}

Beyond generating better insights, \modelname~can also generate more accurate, diverse, and interpretable plots. Using the same incidents management task, we conduct a comparative case study on plot generation between our framework~\modelname~and the best baseline model AgentPoirot. As shown in Table~\ref{tab:case_study_plot}, \modelname~outperforms the baseline in all of the following four aspects.

\noindent\textbf{Richness of Information.}\quad
The plots generated by the baseline model often remain narrowly focused, lacking contextual comparisons or layered breakdowns. For example, when visualizing caller-level trends, the baseline displays isolated line charts without integrating group-level context. \modelname~enriches these visuals by embedding comparative dimensions, such as group-to-individual contrasts, highlighting both unique and systemic patterns. This allows decision-makers to distinguish localized anomalies from broader shifts, directly enhancing strategic relevance.

\noindent\textbf{Diversity and Non-Redundancy.}\quad
As illustrated by the cases in the first and second rows of Table~\ref{tab:case_study_plot}, baseline outputs tend to repeat similar plots across different insights, limiting analytical novelty and depth. In contrast, \modelname~generates diverse and non-redundant plots. This variety not only prevents redundancy but also enhances the diversity and depth of analysis.

\noindent\textbf{Error-Free Code and Plots.}\quad
In the baseline outputs, code logic errors result in misleading plots, such as the bar chart where all bars have identical heights regardless of the data. Through our carefully designed multi-path reasoning and code fix process, \modelname~resolves such issues, producing error-free plots that reflect the true data distribution. It ensures that visualizations can be trusted for decision-making with less manual validation.

\noindent\textbf{Readability and Interpretability.}\quad
The baseline model frequently produces plots with visual clutter, label overlap, and mislabeling. In contrast, \modelname~replaces these with clean, stacked-area or grouped-bar designs, improved color palettes, and simplified axis labeling (e.g., monthly labels ``Jan-Dec’’ instead of raw date strings). \modelname~significantly improves the readability and interpretability of the plots, even for complex multi-category data.
It not only enhances the accuracy of insight generation by the \textit{interpreter} agent but also facilitates human analysts in understanding complex patterns and making informed decisions more effectively, increasing the overall utility and trustworthiness of our framwork.

In summary, while the baseline model can produce basic visual plots, it lacks the accuracy, diversity, and interpretability, especially required for high-stakes operational analysis. \modelname~delivers richer, more varied, and more readable visual plots, transforming plots from passive illustrations into active decision-support tools.

\section{Prompts for Agents}
\label{app:prompt}
\renewcommand\lstlistingname{Prompt}

Prompt~\ref{prompt:search_judge} - Prompt~\ref{prompt:summary} present the detailed prompts for agents in~\modelname.

\onecolumn

\begin{lstlisting}[breaklines=true,label={prompt:search_judge},caption={Prompt for the \textit{judge} agent in RAKG Module.}]
You are a reasoning-enhanced data analysis agent. Your task is to help extract meaningful insights from real-world data.

## Context:
- The analysis goal is: <goal>{goal}</goal>
- The database schema is as follows:
<schema>{schema}</schema>
- Sample data or description of contents:
<info>{description}</info>

## Your Task:
1. Carefully examine the above context and determine whether the analysis can be completed **purely based on the given data**.
2. If not, **identify what kind of external knowledge** (e.g., domain-specific definitions and knowledge, policy changes, socioeconomic context, product classification standards, etc.) is likely needed.
3. Clearly state whether external search is needed:
   - If YES:
     - Describe **why** the search is necessary.
     - Specify **what kind of knowledge** should be retrieved.
   - If NO:
     - Explain why the current information is sufficient.

## Constraints:
- Be cautious not to assume unavailable data.
- Use chain-of-thought reasoning: think step by step before deciding.

## Respond in the following format (If YES, Knowledge Needed):
<reason>Step-by-Step Reasoning</reason>
<answer>YES/NO</answer>
\end{lstlisting}

\begin{lstlisting}[breaklines=true,label={prompt:query_generator},caption={Prompt for the \textit{data-aware query generator} agent in RAKG Module.}]
You are a domain-aware data analyst assistant.

## Context:
- The analysis goal is: <goal>{goal}</goal>
- The database schema is as follows:
<schema>{schema}</schema>
- Sample data or description of contents:
<info>{description}</info>

## Your Task:
1. **Write a list of concise, high-quality and effective Google search queries** to retrieve reliable useful domain knowledge.
2. The query should be focused, unambiguous, and ready for direct use in a Google search engine.
3. You can write at most **{max_queries}** queries. Ensure the richness of multiple queries and avoid semantic duplication.

## Respond in the following format:
<query>The search query</query>
\end{lstlisting}
\pagebreak

\begin{lstlisting}[breaklines=true,label={prompt:knowledge_generator},caption={Prompt for the \textit{knowledge generator} agent in RAKG Module.}]
You are a domain-aware, search-augmented data analysis assistant.

## Objective:
Given a data analysis context and the corresponding search engine results, identify and extract **useful domain knowledge or external facts** that can help guide, validate, or enrich the analysis process.

## Input:
### 1. Data Analysis Context
- The analysis goal is: <goal>{goal}</goal>
- The database schema is as follows:
<schema>{schema}</schema>
- Sample data or description of contents:
<info>{description}</info>

### 2. Search Results
<search_results>
{search_results}
</search_results>

## Task:
* Based on the data analysis context and search results, what additional vertical field and domain knowledge does the analyst need to know to achieve the goal?
* Provide a list of knowledge item that can be used by the data scientists in your team to explore my data and reach my goal.
* Explore diverse aspects of the data, and provide knowledge that is relevant to the analysis goal.
* Do not number the knowledge item.
* Most importantly, each knowledge item must be enclosed within <knowledge></knowledge> tags.

## Output Format:
<knowledge>A knowledge item.</knowledge>
\end{lstlisting}

\begin{lstlisting}[breaklines=true,label={prompt:role_designer},caption={Prompt for the \textit{role designer} agent in Question Raising Module.}]
You are an expert system designer working on a multi-agent data analysis framework. Your current task is to design a **diverse set of problem-posing roles** for the system's "question generation module," where each role adopts a unique perspective to ask insightful questions about a given dataset.

Your goal is to **maximize diversity and coverage** in the types of questions that could uncover valuable, possibly hidden, insights from the data.   Each role should have a clearly defined **perspective**, **focus area**, etc.

### Please output a structured list of roles with the necessary information such as: Role Name, Background and Perspective (e.g., behavioral analyst, business strategist, anomaly detector), Capability, Knowledge, Personality Traits, etc.

### Data Context:

The analysis goal is:
<goal>{goal}</goal>

The schema of the dataset is:
<schema>{schema}</schema>

The detailed information of the dataset file is:
<info>{description}</info>

### Instructions:
* Write a list of roles with the required information.
* You can produce at most {max_roles} roles. Therefore, you need to make a careful choice before answering.
* Most importantly, each role description must be enclosed within <role></role> tags. For example: <role>Description of role1</role> <role>Description of role2</role> ...
\end{lstlisting}
\pagebreak

\begin{lstlisting}[breaklines=true,label={prompt:question_raising},caption={Prompt for \textit{question raising} for different analytical roles in Question Raising Module.}]
[Role Description]
You are:
{role_description}


[Task Description]
I require the services of your team to help me reach my goal.

<goal>{goal}</goal>

<schema>{schema}</schema>

<info>{description}</info>

### Previous Answered Questions:
<prev_questions>{prev_questions}</prev_questions>

Given relevant domain knowledge that may be useful:
{knowledge}

Instructions:
* Produce a list of follow up questions to explore my data and reach my goal.
* Note that we have already answered <question> and have the answer at <answer>, do not include a question similar to the one above. 
* Explore diverse aspects of the data, and ask questions that are relevant to my goal.
* You must ask the right questions to surface anything interesting (trends, anomalies, etc.)
* Make sure these can realistically be answered based on the data schema.
* The insights that your team will extract will be used to generate a report.
* Each question that you produce must be enclosed in <question>content</question> tags.
* Each question should only have one part, that is a single '?' at the end which only require a single answer.
* Do not number the questions.
* You can produce at most {max_questions} questions.
\end{lstlisting}
\pagebreak

\begin{lstlisting}[breaklines=true,label={prompt:question_judge},caption={Prompt for the \textit{judge} agent in Question Raising Module.}]
You are an expert data analyst tasked with selecting the most promising questions from a set of diverse questions proposed by multiple agents, each with a unique role and perspective. Your goal is to identify the questions that are most likely to uncover meaningful, non-trivial insights from the data, but Do Not select a question similar to the previous answered questions. 
These questions may differ in form (e.g., comparative, temporal, behavioral, causal), but your selection should prioritize:
* Potential to reveal non-obvious patterns, relationships, or conclusion
* Relevance to the data schema and business context
* Feasibility of being answered with the given data
* Diversity in angles or dimensions (not redundant)

### Input:
You will be given:
* The number of questions to select: {number}
* The data context and schema
* A list of questions that are answered previously.
* A list of questions proposed by multiple roles.

### Output Requirements:
Please return the following:
1. A list of the {number} selected questions, ranked if possible
2. For each selected question, include:
    * The index of the question
    * The question text
    * The reason for selection (e.g., why it is likely to yield a valuable insight)
    * Output the index of the question in your response inside <question_id></question_id> tag.

### Data Context:
The analysis goal is:
<goal>{goal}</goal>

The schema of the dataset is:
<schema>{schema}</schema>

The detailed information of the dataset file is:
<info>{description}</info>

### Previous Answered Questions:
<prev_questions>{prev_questions}</prev_questions>

### Proposed Questions:
<proposed_questions>{proposed_questions}</proposed_questions>

### Example Output Format:
<question_id>0</question_id> <question>The text of question 0.</question> <reason>The reason for selection.</reason>
<question_id>3</question_id> <question>The text of question 3.</question> <reason>The reason for selection.</reason> ...
Most importantly, the index of questions start with 0, so the selected question index should be between 0-{ques_num} !
\end{lstlisting}
\pagebreak

\begin{lstlisting}[breaklines=true,label={prompt:question_rewriter},caption={Prompt for the \textit{schema-aware question rewriter} agent in Insights Generation Module.}]
[Role Instruction]
You are a schema-aware question reformulator with expertise in database systems.  Your task is to rewrite the given question by explicitly incorporating missing semantic information of the datasets, deconstruct ambiguities, clarify objectives to maximize clarity and actionability of the question. Follow these steps strictly:
1.  Intent Deconstruction
# Extract the core verb phrase (VP) and key named entities (NE) from the original question.
# Identify ambiguous or incomplete semantics due to missing schema elements.

2. Semantic Anchoring
# Enhance the question by explicitly adding:
a) Missing table/column names (as indicated in the schema&information of the datasets)
b) Temporal constraints (e.g., semester, year, quarter) if relevant
c) Categorical dimensions (e.g., student type, degree level)
d) Aggregation requirements (e.g., sum, average, count)

3. Deconstruct Ambiguities and Clarify Objectives
# Identify, re-express or explain vague terms, implicit assumptions, acronyms or undefined variables in the original question.
# Extract the core intent (What must the answer achieve?)  

4.  Structural Reformulation
# Restructure the question. Format as self-contained, clear query with necessary context embedded.

[Output Requirements]
# Preserve all technical terms from the original question.
# Do not include explanatory notes or unrelated content.
# Most importantly, the rewritten question must be enclosed within <question></question> tags. Refer to the example response below.

[Example Demonstration]
Question: Find the **avg** age of faculty members.
Schema: [["name": faculty_birth_year, "type": object, "missing_count": 0, "unique_count": 124], ...]
Information: The detailed information of the dataset.
Rewritten Question: <question>In a database containing faculty_birth_year and department_info, how to calculate the average age of faculty members grouped by department_name?</question>

[Task Execution]
Now, rewrite the following question based on the provided schema&information of the datasets:
Question:
{question}

Schema:
{schema}

Information:
{description}

Domain knowledge:
{knowledge}

Rewritten Question:
\end{lstlisting}
\pagebreak

\begin{lstlisting}[breaklines=true,label={prompt:code_generation},caption={Prompt for \textit{multi-path code generation} in Insights Generation Module.}]
SYSTEM PROMPT for Divide-and-Conquer:
    You are a professional database administrator and expert software engineer specializing in code writing and improvement. You should always reason with a "divide-and-conquer" mindset before writing the code: Decompose the user's question into the smallest independent sub-tasks possible; Solve the sub-tasks sequentially, making your intermediate reasoning explicit; You must output your detailed step-by-step "divide-and-conquer" reasoning within the tag <reasoning></reasoning>; You must comment all your reasoning process in the code using '#', including the content within <reasoning></reasoning> tags! ; Only after the reasoning is complete, write the final code (faithful to the reasoning).

SYSTEM PROMPT for Query Plan:
    You are a professional database administrator and expert software engineer specializing in code writing and improvement. Before writing code, produce an explicit **query-plan** in plain English: Outline each logical operation in order-filters, joins, sub-queries / CTEs, aggregations, sorts; Reference exact table / column names from the schema; State the expected intermediate DataFrame or Series after every step; You must output your detailed "query-plan" reasoning within the tag <reasoning></reasoning>; You must comment all your reasoning process in the code using '#', including the content within <reasoning></reasoning> tags! ; Only after the plan is complete, write the final code (faithful to the plan).

SYSTEM PROMPT for Negative Reasoning:
    You are a professional database administrator and expert software engineer specializing in code writing and improvement. You are very meticulous and good at avoiding common pitfalls. Before writing code, produce an explicit **Counterfactual / Negative Reasoning** in plain English: List plausible errors or misconceptions someone might make when answering the question (e.g., double counting, wrong date window, missing NULLs, dtype mismatches); For each risk, explain how your Python code will prevent it; You must output your detailed "Counterfactual / Negative Reasoning" within the tag <reasoning></reasoning>; You must comment all your reasoning process in the code using '#', including the content within <reasoning></reasoning> tags! ; Only after the reasoning is complete, write the final code (faithful to the reasoning).
\end{lstlisting}

\begin{lstlisting}[breaklines=true,label={prompt:code_selector},caption={Prompt for the \textit{code selector} agent in Insights Generation Module.}]
You are a kind, experienced and professional database administrator and code auditor tasked with selecting the correct code from a set of diverse code written by multiple agents, each with a unique coding perspective.


### Input:
You will be given:
* The data context, schema and the user question.
* A list of codes trying to answer the question.
* Other Code Requirements.


### Output Requirements:
Please return the following:
* The index of the selected code.
* The reason for selection.
* Output the index of the code in your response inside <code_id></code_id> tag.


### Data Context:
The analysis goal is:
<goal>{goal}</goal>

The user question is:
<question>{question}</question>

The schema of the dataset is:
<schema>{schema}</schema>

The detailed information of the dataset file is:
<info>{description}</info>

The path of the dataset file is:
<path>{database_path}<path>

Other Code Requirements are:
<requirements>
* Make a single code block for starting with ```python
* Do not produce code blocks for languages other than Python.
* Import pandas as pd, matplotlib.pyplot as plt, numpy as np, ...
* You may use the predefined functions mentioned above to generate plots, or write your own plotting functions. If you choose not to use the predefined functions, you must save the generated plot using the following code: plt.savefig("plot.jpg") plt.close()
* You must generate one single simple plot and save it as a jpg file.
* The generated plot will be analyzed by a data analyst, so please ensure it is as clear and easy to understand as possible.
* Plot Best Practices:
  - Adds appropriate titles, labels, and annotations that highlight the key insights.
  - For legends: Always use clear, descriptive legend titles and place them optimally (usually upper right or outside).
  - For color selection: Use colorblind-friendly palettes (viridis, plasma, cividis) or plt.cm.Paired
  - For multiple series: When plotting multiple data series, either: Use plt.subplots to create separate plots, or Use proper stacking techniques with stacked=True parameter. Avoid overwriting plots on the same axes unless showing direct comparisons.
  - For pie charts: Use plt.axis('equal') to ensure proper circular appearance.
  - For data preparation: Prepare data properly before visualization (aggregation, transformation). For example, use pandas aggregation (crosstab, pivot_table) before plotting.
  - For formatting: Include appropriate sizing and formatting for all visual elements. For example, set appropriate fontsize for title (14), labels (12), and tick labels (10). Calls plt.tight_layout() if necessary.
* For the plot, save a stats json file that stores the data of the plot.
* For the plot, save a x_axis.json and y_axis.json file that stores 100 most important x and y axis data points of the plot, respectively.
* For the json file must have a "name", "description", and "value" field that describes the data.
* If the content of the json file is getting too long, truncate the unnecessary parts until the number of characters is less than 10000.
* End your code with ```.
</requirements>


### Candidate code:
## Code Candidate 0:
<code_0>
{code_0}
</code_0>
## Code Candidate 1:
<code_1>
{code_1}
</code_1>
## Code Candidate 2:
<code_2>
{code_2}
</code_2>


### Example Output Format:
<code_id>5</code_id><reason>The reason for selection.</reason>
Most importantly, the index of code candidate start with 0, so the selected code index should be between 0-2 !

### Response:
\end{lstlisting}
\pagebreak

\begin{lstlisting}[breaklines=true,label={prompt:code_reviewer},caption={Prompt for the \textit{code reviewer} agent in Insights Generation Module.}]
You are a kind, experienced and professional database administrator and code auditor. Given the [Database schema] descriptions, [Database detailed information] descriptions, and a user [Question] with corresponding [Python code], you need to analyze the provided code. Start from the components of [Python code], step by step, to analyze whether it aligns with the user [Question] intent, aligns with the [Database schema], and any possible errors occur.

There are some criteria:
### 1.  Requirement Alignment
- Verify if the code fully addresses the user's original question
- Identify discrepancies between requested functionality and implementation

### 2.  Database Compliance
**Schema Validation**:
- Check table/column naming against schema definitions
- Validate data types (e.g., VARCHAR length vs actual data)
- Verify constraints (PK/FK/UNIQUE/NOT NULL) are properly handled

**Data Format Integrity**:
- Ensure date/numeric formats match database settings
- Confirm encoding/charset consistency
- Check bulk operation compatibility (batch size, transaction scope)

### 3.  Problem Detection
**Critical Issues**:
- SQL injection vulnerabilities
- Race conditions in concurrent operations
- Resource leaks (connections, cursors)

**Operational Risks**:
- N+1 query patterns
- Missing transaction boundaries
- Index misuse (e.g., non-sargable queries)

**Data Integrity Concerns**:
- Dirty read/write scenarios
- Improper null handling
- Silent truncation risks

### 4.  Output Format Requirements
**Structured Response Requirements**

[Goal]
{goal}
[Database schema]
{schema}
[Database detailed information]
{description}
[Database path]
{database_path}
[Question]
{question}
[Other Code Requirements]
* Make a single code block for starting with ```python
* Do not produce code blocks for languages other than Python.
* Import pandas as pd, matplotlib.pyplot as plt, numpy as np, ...
* You may use the predefined functions mentioned above to generate plots, or write your own plotting functions. If you choose not to use the predefined functions, you must save the generated plot using the following code: plt.savefig("plot.jpg") plt.close()
* You must generate one single simple plot and save it as a jpg file.
* The generated plot will be analyzed by a data analyst, so please ensure it is as clear and easy to understand as possible.
* Plot Best Practices:
  - Adds appropriate titles, labels, and annotations that highlight the key insights.
  - For legends: Always use clear, descriptive legend titles and place them optimally (usually upper right or outside).
  - For color selection: Use colorblind-friendly palettes (viridis, plasma, cividis) or plt.cm.Paired
  - For multiple series: When plotting multiple data series, either: Use plt.subplots to create separate plots, or Use proper stacking techniques with stacked=True parameter. Avoid overwriting plots on the same axes unless showing direct comparisons.
  - For pie charts: Use plt.axis('equal') to ensure proper circular appearance.
  - For data preparation: Prepare data properly before visualization (aggregation, transformation). For example, use pandas aggregation (crosstab, pivot_table) before plotting.
  - For formatting: Include appropriate sizing and formatting for all visual elements. For example, set appropriate fontsize for title (14), labels (12), and tick labels (10). Calls plt.tight_layout() if necessary.
* For the plot, save a stats json file that stores the data of the plot.
* For the plot, save a x_axis.json and y_axis.json file that stores 100 most important x and y axis data points of the plot, respectively.
* For the json file must have a "name", "description", and "value" field that describes the data.
* If the content of the json file is getting too long, truncate the unnecessary parts until the number of characters is less than 10000.
* End your code with ```.
[Python code]
<code>
{code}
</code>


Does the [Python code] has errors? If Yes, provide detailed explanations and reviews for subsequent modifications.

* You must output the summary review of the code in your response inside <review></review> tag.
* You only need to point out the mistakes in the code and don't point out the correct things !!!
* You must focus on logical accuracy and functional requirements. Ignore minor issues unless they impact functionality.
* You must output the final judgment ('Yes' or 'No') in your response inside <judgment></judgment> tag. 'Yes' means there are problems with the code.

### Your answer:
\end{lstlisting}
\pagebreak

\begin{lstlisting}[breaklines=true,label={prompt:plot_reviewer},caption={Prompt for the \textit{plot reviewer} agent in Insights Generation Module.}]
You are a highly skilled data visualization evaluator and database administrator tasked with evaluating a data visualization result. Your task is to assess a Python-generated data plot based on the user's natural language question, the underlying database schema or description, the code used to generate the plot, and the plot image itself.


## Given the following inputs:
### 1.  Data Analysis Context
- The analysis goal is: <goal>{goal}</goal>
- The analysis question is: <question>{question}</question>
- The database schema is as follows:
<schema>{schema}</schema>
- Sample data or description of contents:
<info>{description}</info>
- The data path of the database:
<data_path>{database_path}</data_path>

### 2.  Visualization Results
- The current Python code that generates the plot.
<code>
{code}
</code>
- The list of predefined functions that may be used in the code and their example usage:
<function>
{function_docs}
</function>
- The generated plot image in 'input_image' (if available).


## Instructions:
* Determine whether the generated plot accurately reflects the user question.
* Identify mismatches between the question and the plot, or between the plot and the database.  If the plot is missing, analyze the code to infer the failure reason.
* Suggest clear, actionable, and detailed instructions to fix issues and improve the visualization.
* Propose improvements focused on clarity, readability, correct labeling, axis formatting, legends, colors, and any other visual elements that would make the plot more informative and aligned with the user's intent.
* If the code need to be fixed, provide detailed explanations and reviews for subsequent modifications.
* Don't be too critic!!! Ignore minor issues unless they do have a significant impact.
* You only need to point out the issues and don't point out the correct things !!!


## Output Requirements
* You must output the final judgment ('Yes' or 'No') in your response inside <judgment></judgment> tag.  'Yes' means there are problems with the code.
* You must output the summary review or instructions in your response inside <review></review> tag.
\end{lstlisting}

\begin{lstlisting}[breaklines=true,label={prompt:code_fixer},caption={Prompt for the \textit{code fixer} agent in Insights Generation Module.}]
You are a professional database administrator and expert software engineer specializing in code writing and improvement.

Given the goal:
{goal}

Given the schema:
{schema}

Given the information of the dataset file:
{description}

Given the data path:
{database_path}

Given the list of predefined functions and their example usage:
{function_docs}

A programmer has written the python code required to answer this user question "{question}" but the code reviewer and the plot reviewer gave some reviews. Your task is to refactor or rewrite code based on provided review feedback. Your code must align with the user question, align with the dataset schema & format, and avoid introducing additional errors.

The review of the code reviewer is:
{review}

The review of the plot reviewer is:
{plot_review}

The original code is:
<code>
{code}
</code>

You should:
* Make a single code block for starting with ```python
* Do not produce code blocks for languages other than Python.
* Import pandas as pd, matplotlib.pyplot as plt, numpy as np, ...
* You may use the predefined functions mentioned above to generate plots, or write your own plotting functions. If you choose not to use the predefined functions, you must save the generated plot using the following code: plt.savefig("plot.jpg") plt.close()
* You must generate one single simple plot and save it as a jpg file.
* The generated plot will be analyzed by a data analyst, so please ensure it is as clear and easy to understand as possible.
* Plot Best Practices:
  - Adds appropriate titles, labels, and annotations that highlight the key insights.
  - For legends: Always use clear, descriptive legend titles and place them optimally (usually upper right or outside).
  - For color selection: Use colorblind-friendly palettes (viridis, plasma, cividis) or plt.cm.Paired
  - For multiple series: When plotting multiple data series, either: Use plt.subplots to create separate plots, or Use proper stacking techniques with stacked=True parameter. Avoid overwriting plots on the same axes unless showing direct comparisons.
  - For pie charts: Use plt.axis('equal') to ensure proper circular appearance.
  - For data preparation: Prepare data properly before visualization (aggregation, transformation). For example, use pandas aggregation (crosstab, pivot_table) before plotting.
  - For formatting: Include appropriate sizing and formatting for all visual elements. For example, set appropriate fontsize for title (14), labels (12), and tick labels (10). Calls plt.tight_layout() if necessary.
* For the plot, save a stats json file that stores the data of the plot.
* For the plot, save a x_axis.json and y_axis.json file that stores 100 most important x and y axis data points of the plot, respectively.
* For the json file must have a "name", "description", and "value" field that describes the data.
* If the content of the json file is getting too long, truncate the unnecessary parts until the number of characters is less than 10000.
* End your code with ```.

Output code:
\end{lstlisting}
\pagebreak

\begin{lstlisting}[breaklines=true,label={prompt:interpreter},caption={Prompt for the \textit{interpreter} agent in Insights Generation Module.}]
### Instruction:
You are trying to answer a question based on information provided by a data scientist.

Given the following dataset schema:
<schema>{schema}</schema>

Given the following dataset information:
<info>{description}</info>

Given the goal:
<goal>{goal}</goal>

Given the question:
<question>{question}</question>

Give the analysis code:
<code>
{code}
</code>

Given the list of predefined functions and their example usage:
{function_docs}

Given the analysis results:
<analysis>
    <message>
        {message}
    </message>
    {insights}
</analysis>

We also provide the plot generated by the code in 'input_image'.

Instructions:
* Based on the code, analysis, plot and other information provided above, write an answer to the question enclosed with <question></question> tags.
* The answer should be a single sentence, but it should not be too high level and should include the key details from justification.
* Write your answer in HTML-like tags, enclosing the answer between <answer></answer> tags, followed by a justification between <justification></justification> tags, followed by an insight between <insight></insight> tags.
* Refer to the following example response for the format of the answer and justification.
* The insight should be something interesting and grounded based on the question, goal, and the dataset schema, something that would be interesting. 
* The insight should be as quantiative as possible and informative and non-trivial and concise.
* The insight should be a meaningful conclusion that can be acquired from the analysis in laymans terms

Example response:
<answer>This is a sample answer</answer>
<insight>This is a sample insight</insight>
<justification>This is a sample justification</justification>

### Response:
\end{lstlisting}
\pagebreak

\begin{lstlisting}[breaklines=true,label={prompt:insight_judge},caption={Prompt for the \textit{final judge} agent in Insights Generation Module.}]
You are a reasoning-enhanced data analysis assistant.


## Objective:
Given a data analysis context, and multiple responses from different data analysis assistant to a data analysis question. Each response contains three parts: <answer>, <insight>, and <justification>. Your task is to consider all given responses and background information to produce a final, well-reasoned results. You must understand the context, evaluate the consistency and quality of the individual responses, and summarize the most credible and insightful points.


## Input:
### 1. Data Analysis Context
- The analysis goal is: <goal>{goal}</goal>
- The database schema is as follows:
<schema>{schema}</schema>
- Sample data or description of contents:
<info>{description}</info>

### 2. The data analysis question
<question>{question}</question>

### 3. Multiple responses from different data analysis assistant
<responses>
{responses}
<responses>


## Requirements:
- Use all available context to reason.
- Your output must also include exactly the following tags:
  <answer>...</answer>
  <insight>...</insight>
  <justification>...</justification>
- The answer should be a single sentence, but it should not be too high level and should include the key details from justification.
- The insight should be something interesting and grounded based on the question, goal, and the dataset schema, something that would be interesting.
- The insight should be as quantitive as possible and informative and non-trivial and concise.
- The insight should be a meaningful conclusion that can be acquired from the analysis in layman terms.
- Justification should clearly explain why the synthesized result is valid, including reasoning over conflicting or supporting evidence from different replies.


## Output Format:
<answer>[Your final answer]</answer>
<insight>[Your refined insight derived from all responses]</insight>
<justification>[Detailed reasoning justifying why this synthesis is the best possible]</justification>
\end{lstlisting}
\pagebreak

\begin{lstlisting}[breaklines=true,label={prompt:summary},caption={Prompt for \textit{insights summarizing}.}]
Hi, I require the services of your team to help me reach my goal.

<context>{context}</context>

<goal>{goal}</goal>

<history>{history}</history>

Instructions:
* Given a context and a goal, and all the history of <question_i><answer_i> pairs from the above list, generate the 3 top actionable insights.
* Make sure they don't offer actions and the summary should be more about highlights of the findings.
* Output each insight within this tag <insight></insight>.
* Each insight should be a meaningful conclusion that can be acquired from the analysis in layman terms and should be as quantitative as possible and should aggregate the findings.
\end{lstlisting}

\end{document}